\documentclass{article}

\usepackage{microtype}
\usepackage{graphicx}
\usepackage{booktabs} 

\usepackage{hyperref}

\usepackage[accepted]{icml2024}

\usepackage{amsmath}
\usepackage{amssymb}
\usepackage{mathtools}
\usepackage{amsthm}

\usepackage[capitalize,noabbrev]{cleveref}

\theoremstyle{plain}
\newtheorem{theorem}{Theorem}[section]

\newtheorem{lemma}[theorem]{Lemma}

\theoremstyle{definition}
\newtheorem{definition}[theorem]{Definition}
\newtheorem{assumption}[theorem]{Assumption}

\usepackage[textsize=tiny]{todonotes}

\newcommand{\vx}{\mathbf{x}}
\usepackage{subdef}
\usepackage[textsize=tiny]{todonotes}
\usepackage{amsmath}
\usepackage{multirow}
\def\rot{\rotatebox}
\usepackage{todonotes}
\RequirePackage[inline,shortlabels]{enumitem}
\setlist{nosep}

\usepackage[subtle]{savetrees}

\usepackage{algorithm}

\newcommand{\swap}[3][-]{312} 

\newcommand{\xhdr}[1]{\vspace{0.0mm}\noindent{\bf #1.}}
\renewcommand{\beta}{t}
\renewcommand{\rho}{y}

\newcommand{\phinet}{{\hat{\tau}}}

\newcommand{\our}{PairNet}
\newcommand{\factual}{Factual}

\newcommand{\cX}{\mathcal{X}}
\newcommand{\cK}{\mathcal{K}}
\newcommand{\cT}{\mathcal{T}}
\newcommand{\cY}{\mathbb{R}}
\newcommand{\cN}{\mathcal{N}}
\newcommand{\cNN}{\text{Nbr}}
\newcommand{\distance}{d}
\renewcommand{\algorithmiccomment}[1]{\hfill$\triangleright$\textit{\color{blue} \footnotesize{{#1}}}}

\usepackage{graphicx}
\usepackage{subcaption}
\usepackage{booktabs} 
\usepackage{xcolor,colortbl}
\definecolor{green}{rgb}{0.8,1,0.8}
\definecolor{yellow}{rgb}{1,1,0.87}
\definecolor{cyan}{rgb}{0.0, 1.0, 1.0}
\newcommand{\first}[1]{{\cellcolor{green} #1}}
\newcommand{\second}[1]{{\cellcolor{yellow} #1}}
\newcommand{\third}[1]{{\cellcolor{cyan!50} #1}}
\usepackage{soul}
\newcommand{\hlc}[2][yellow]{{%
    \colorlet{foo}{#1}%
    \sethlcolor{foo}\hl{#2}}%
}
\usepackage{amssymb}
\usepackage{pifont}
\newcommand{\cmark}{\ding{51}}%
\newcommand{\xmark}{\ding{55}}%
\usepackage{makecell}
\usepackage{tikz}
\usetikzlibrary{positioning, arrows.meta}

\usepackage{tabularx}
\usepackage{ragged2e}

\newcommand{\embedfct}{\psi}

\makeatletter
\newcommand*{\centernot}{%
  \mathpalette\@centernot
}
\def\@centernot#1#2{%
  \mathrel{%
    \rlap{%
      \settowidth\dimen@{$\m@th#1{#2}$}%
      \kern.5\dimen@
      \settowidth\dimen@{$\m@th#1=$}%
      \kern-.5\dimen@
      $\m@th#1\not$%
    }%
    {#2}%
  }%
}
\makeatother
\newcommand{\independent}{\perp\mkern-9.5mu\perp}
\newcommand{\nindep}{\centernot{\independent}}

\icmltitlerunning{PairNet: Training with Observed Pairs to Estimate Treatment Effect on Individual}
\begin{document}

\setlength{\abovedisplayskip}{0pt}
\setlength{\belowdisplayskip}{0pt}
\setlength{\abovedisplayshortskip}{0pt}
\setlength{\belowdisplayshortskip}{0pt}

\twocolumn[
\icmltitle{PairNet: Training with Observed Pairs to Estimate Individual Treatment Effect}
\icmlsetsymbol{equal}{*}

\begin{icmlauthorlist}
\icmlauthor{Lokesh Nagalapatti}{equal,yyy}
\icmlauthor{Pranava Singhal}{equal,yyy}
\icmlauthor{Avishek Ghosh}{yyy}
\icmlauthor{Sunita Sarawagi}{yyy}
\end{icmlauthorlist}

\icmlaffiliation{yyy}{IIT Bombay}

\icmlcorrespondingauthor{Lokesh Nagalapatti}{nlokeshiisc@gmail.com}
\icmlcorrespondingauthor{Pranava Singhal}{pranava.psinghal@gmail.com}
\icmlcorrespondingauthor{Avishek Ghosh}{avishek\_ghosh@iitb.ac.in}
\icmlcorrespondingauthor{Sunita Sarawagi}{sunita@iitb.ac.in}

\icmlkeywords{Causal Inference, Treatment Efect Estimation, Binary and continuous treatments}

\vskip 0.3in
]



\printAffiliationsAndNotice{\icmlEqualContribution} 

\begin{abstract}
Given a dataset of individuals each described by a covariate vector, a treatment, and an observed outcome on the treatment, the goal of the individual treatment effect (ITE) estimation task is to predict outcome changes resulting from a change in treatment. A fundamental challenge is that in the observational data, a covariate's outcome is observed only under one treatment, whereas we need to infer the difference in outcomes under two different treatments. Several existing approaches address this issue through training with inferred pseudo-outcomes, but their success relies on the quality of these pseudo-outcomes. We propose \our, a novel ITE estimation training strategy that minimizes losses over pairs of examples based on their factual observed outcomes. 
Theoretical analysis for binary treatments 
reveals that \our\ is a consistent estimator of ITE risk, and achieves smaller generalization error than baseline models. Empirical comparison with thirteen existing methods across eight benchmarks, covering both discrete and continuous treatments, shows that \our\ achieves significantly lower ITE error compared to the baselines.  Also, it is model-agnostic and easy to implement. We release the code at the URL: \url{https://github.com/nlokeshiisc/pairnet_release}.
\end{abstract}

\section{Introduction}
Many applications in medicine, finance, and retail require the ability to predict the change in outcome resulting from changing actions or treatments. Classical methods relied on expensive randomized control trial experiments, but with the increasing availability of large observational datasets, a recent research focus is harnessing this observational data to train models for estimating treatment effects. For example, a retail store might have millions of customers whose click-through rates on advertisements have been observed under various settings of treatments such as discounts. Using such datasets it may be possible to train a model to estimate the effect of discounts on click-through rate. However, a challenge is that an individual is observed only under one treatment, and there is no direct supervision of how an individual's outcome could change with changing treatment. Thus, one baseline approach~\cite{cfrnet,vcnet,Tarnet, chauhan2023adversarial, inducbias, overlapping_rep, dragonnet} 
is to train a model to fit outcomes separately for each treatment, and during inference output the difference of predicted outcomes under the two treatments.
Another major line of work is to 
impute pseudo-outcomes for missing treatments in the training dataset and directly supervise outcome differences using them. Imputation methods include meta-learners~\cite{xlearner,rlearner,CurthS23}, matching methods~\cite{matching_survey, prop_score_matching, coarsened_exact_matching, perfect_match, deepmatch}, and generative models such as GANs~\cite{scigan,cevae}.
However, the success of such methods depends on the quality of the inferred pseudo-outcomes. 

We propose an alternative training strategy called \our, that avoids committing to noisy supervision from pseudo-outcomes and instead works only with observed factual outcomes. Unlike all previous methods, \our\ imposes a loss on the \emph{difference of outcomes of pairs of instances}. The pairs are chosen to be close in the covariate space while having different treatments. 
We show that such paired instance-based training more closely aligns with the ITE estimation task than existing methods. 

We theoretically analyze Pair loss for binary treatments and bound the ITE risk. These bounds, expressed in the form of IPM distance between the provided training distribution and a neighborhood distribution, are shown to be tighter when compared to the baseline factual model, which relies on the IPM distance between treated and control distributions. Additionally, we establish that \our\ serves as a consistent estimator of treatment effect.

In summary, we make the following contributions:

\begin{enumerate}[leftmargin=0.2cm, itemsep=0pt]
    \item We introduce \our, a novel approach that {\em only} applies factual losses to observed instance pairs in contrast to several existing methods that infer pseudo-outcomes. \our\ is {\em model-agnostic} and applies to {\em both} discrete and continuous treatments.
    \item  We theoretically show that the ITE risk can be upper bounded by Pair loss and the distributional distance of near neighbours with contrasting treatments. We further show that \our\ is a consistent estimator of ITE under commonly used strict overlap assumption and offers tighter generalization bounds that factual loss.
    \item We compare \our\ with eleven prior methods on three benchmarks for binary treatments and two prior methods on five benchmarks for continuous treatments. We observe that \our\ provides significant gains. 
    We conduct several experiments to explain the superior performance of \our\ and analyze its sensitivity.
\end{enumerate}

\section{Problem Statement}
We follow the Neyman-Rubin potential outcomes framework~\cite{Rubin2005} where an individual with observed covariates $\vx \in \cX$, when subjected to a treatment $t \in \cT$, exhibits an outcome $Y(t)\in \cY$.  The space of treatments $\cT$ could be binary $(\cT = \{0,1\})$ or continuous ($\cT = [0,1]$). 
We are given an observational dataset $D=\{(\vx_i,t_i,y_i):i \in 1\ldots N\}$ where $(\vx,t)$ are samples drawn from a joint distribution $P(X,T)$ such that $X \nindep T$, and $y_i$ is the observed outcome. 
We denote the covariate distribution $P(X = \vx)$ as $p(\vx)$, and the conditional distribution $P(X = \vx | T = t)$ as $p_t(\vx)$. 
The marginal treatment distribution is $P(T = t) = u_t$. 
Let the expected outcome when a covariate $\vx$ is given a treatment $t$ be denoted as  $\mu^*(\vx,t)=\mathbb{E}[(Y(t)|\vx)]$. 
Our goal is to learn a model $\phinet(\vx,t,t')$ that estimates the change in outcome $\tau^*(\vx,t,t')=\mu^*(\vx,t)-\mu^*(\vx,t')$ when a test $(\vx,t)\sim P(X,T)$ is given a new treatment $t'$ that is sampled arbitrarily.  
Solving this problem requires us to minimize the following \textbf{ITE risk}:

\begin{align}
\label{eq:risk}
\mathbb{E}_{P(X,T),T'\ne T} [\|(\tau^*(X,T,T')-\phinet(X,T,T'))\|_2]
\end{align}

\xhdr{Assumptions} Like in prior work we make the following assumptions for identifying ITE from an observational dataset:
\begin{enumerate*}[leftmargin=0.66cm]
\item[A1] \emph{Overlap of treatment:} Every individual has a non-zero probability of being assigned any treatment, i.e., $P(t|\vx) \in (0, 1) ; \forall \vx \in \Xcal, t \in \cT$.
\item [A2] \emph{Consistency:} The observed outcome is the same as the potential outcome, i.e., when an individual is given a treatment $t$, we observe $Y(t)$.
\item[A3] \emph{Unconfoundedness:} The observed covariates $X$ block all backdoor paths between the treatments $T$ and outcomes $Y$.
\end{enumerate*}

\xhdr{Challenges}
The fundamental challenge in estimating $\phinet$ from the observational dataset is that for a given $\vx$, we observe its outcome under only one treatment sampled according to $P(t|\vx) = \pi_t(\vx)$.  Thus, we cannot directly supervise $\phinet(\vx,t,t')$ using $D$. A common practice is to estimate $\hat{\mu}(\vx,t)$ instead by minimizing the following \textbf{factual} loss:
\begin{align}
\label{eq:floss}
    \hat{\mu} = \argmin_\mu \sum_{i=1}^N (\rho_i-\mu(\vx_i,\beta_i))^2
\end{align}
and then estimate the treatment effect as 
\begin{align}
\label{eq:floss:ite}
\phinet(\vx,\beta,\beta') = \hat{\mu}(\vx,\beta)-\hat{\mu}(\vx,\beta')
\end{align}
We call this the \textbf{\factual\ model}.
The main limitation of the factual loss-based training is that the loss is independent for each treatment whereas, for ITE risk, we wish to minimize error in the difference in outcomes of two treatments $t,t'$.  The gap between factual risk and ITE risk gets even worse because of confounding ($T \nindep X$) causing $p_t(\vx) \ne p_{t'}(\vx)$ in general. For example, for binary treatment, the $\hat{\mu}(\vx,1)$ trained on examples sampled from $p_1(\vx)$, may perform poorly when during inference it is deployed for treatment effect estimation on $\vx \sim p_0(\vx)$.

\section{Related Work}
\label{sec:relate}
\vspace{-1mm}
We categorize the extensive literature on ITE estimation based on whether they train with pseudo-outcomes, or not. 
\vspace{-2.5mm}
\subsection{Training with pseudo-outcomes}
\vspace{-1mm}
A common strategy to address the ITE task is to estimate the pseudo-outcomes for missing treatments and then use them to minimize the ITE risk in Eq. \ref{eq:risk}. These methods differ primarily in the method in which the pseudo-outcomes are inferred.

\xhdr{Meta Learners} 
One class of approaches directly models the treatment effect $\hat{\tau}$ under the assumption that the treatment effect function $\tau$ is simpler than the individual potential outcome functions $\mu_t$~\cite{Gao2020,inducbias}. 
A popular strategy is using two-stage meta-learning. First, train a \factual\ model $\hat{\mu}(\vx,t)$ to estimate outcome using $D$ as shown in Eq~\ref{eq:floss}, and then train the ITE model $\hat{\tau}(\vx,t,t')$ using pseudo ITE $y_i(t_i)-\hat{\mu}(\vx_i,t')$.
X-Learner~\cite{xlearner} is one of the earliest such meta-learners.
DR-Learner \cite{drlearner} is similar but adds a propensity score model in the first stage to synthesize doubly robust ITE labels $\hat{\tau}$. R-Learner\cite{rlearner} employs Robinson's error decomposition to directly model $\tau$.

\xhdr{Matching Methods}
Matching methods \cite{matching_survey, prop_score_matching, coarsened_exact_matching, perfect_match, deepmatch}
impose losses on missing treatments by borrowing outcomes from near neighbors in the dataset. They use techniques like propensity scores, exact matching, stratification, covariate distance metrics, etc., to form neighbors across treatment groups. 
However, if the distance metric connecting neighbors is flawed, the inferred outcomes can be wrong leading to faulty supervision. 
While \our\ also creates neighbors, it imposes losses solely using the observed outcomes in the dataset, and thus the accuracy of \our\ is unaffected by the correctness of any pseudo outcomes.

\xhdr{Generative methods}
GANITE \cite{ganite} uses GANs to synthesize pseudo-outcomes and uses them to learn $\tau$.   SciGAN \cite{scigan} extends GANITE \cite{ganite} for continuous treatments. Gaussian Processes are employed in \cite{alaa2017bayesian, alaa2017deep,  overlapping_rep} and Variational Autoencoders (VAEs) are explored in \cite{cevae, cevae_critical, reconsidering_gen}. 
\nocite{tree_1, tree_2}

\vspace{-2mm}
\subsection{Training without Pseudo-outcomes}
\label{sec:relate:cont}
\vspace{-1mm}
These methods address the treatment confounding challenge by learning shared representations with regularizers to correct the effect of confounding. 
A popular method is TARNet \cite{Tarnet} which uses a shared representation layer and defines separate prediction heads for each treatment group. 
CFRNet \cite{cfrnet} additionally applies IPM regularizers to balance representations. \citet{replearning_ite} encourages representation similarity based on propensity scores. FlexTENet \cite{inducbias} introduces further sharing of parameters across treatment-specific layers.
DragonNet \cite{dragonnet} introduces Targeted Regularizers for CATE estimation. 
\textbf{Weighting methods:} Certain methods account for confounding by imposing weighted factual losses \cite{ctr_importance, disentangled_ite, weighted_factual, weighted_factual_2, dragonnet}. DragonNet \cite{dragonnet} is a doubly robust method that implements the Augmented Inverse Probability Weighted estimator. But these methods depend on propensity scores which are often not well-calibrated. 

\xhdr{Continuous Treatments Effects}
DRNet \cite{DRNet} first handled continuous treatments using a TARNet-like architecture with binned treatments. VCNet uses spline basis expansion for $t$, with regularizers as either Targeted\cite{vcnet} or the Hilbert Schmidt Independence Criterion (HSIC) \cite{vcnet_hsic}. GIKS \cite{giks} uses data augmentation to break the confounding in the training dataset. \citet{TransTEE} proposes a transformer-based model designed specifically for text data.



\section{The Pair Loss} 
\label{sec:pairnet}
Unlike existing approaches, that attempt to impute outcomes of missing treatment $t'$, \our\ minimizes the difference in observed outcomes of two individuals.  The pairs are chosen to be close in the covariate space and are observed under different treatments.  Specifically, they are created as follows:  For each instance $(\vx_i,\beta_i, y_i) \in D$, sample an alternative treatment $\beta'$, find instance subsets from $D$ to form neighbor set $\cNN(\vx_i,\beta') = \{(\vx_j, \beta_j)\}$ such that $\beta_j \approx \beta'$ and distance $d(\vx_i, \vx_j) = ||\embedfct(\vx_i)- \embedfct(\vx_j)||$ is small. Here, $\embedfct$ denotes any embedding function that can be used to compute distances. This sampling procedure induces a distribution over the neighbours $(\vx', \beta')$ for observed $(\vx, \beta)$ which we denote as $q_t(\vx', \beta' | \vx) \propto e^{-d(\vx, \vx')}p_{t'}(\vx')$. \our\ models potential outcomes as $\hat{\mu}(\vx, \beta) = \mu(\phi(\vx), \beta)$, similar to representation learning methods where $\phi$ is a representation extraction function shared across treatments, and $\mu$ is the treatment-specific outcome prediction function. It imposes a loss on a pair of instances as follows:
\begin{equation}
\label{eq:sAgree}
\begin{split}
  \hat{\mu}(\vx,\beta)  = \argmin_{\mu,\phi}\sum_{i=1}^N \sum_{\beta' \ne t_i}\sum_{(\vx_i', t_i')\in \cNN(\vx_i,\beta')} ((\rho_i - \rho'_i) \\  
  - (\mu(\phi(\vx_i),\beta_i)-\mu(\phi(\vx'_i),\beta'_i)))^2 
\end{split}
\end{equation}

\begin{figure}
    \centering
    \begin{subfigure}{0.24\textwidth}
        \centering
        \includegraphics[width=\textwidth]{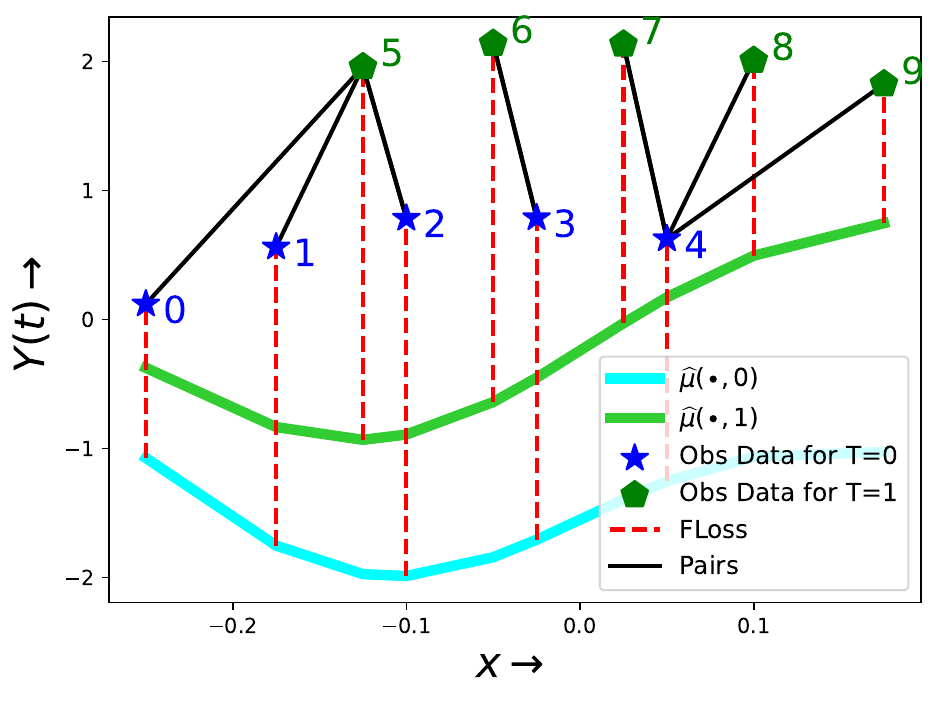} 
        \caption{Ground truth vs. predicted $\hat{\mu}$ functions}
        \label{fig:mot}
    \end{subfigure}%
    \hfill
    \begin{subfigure}{0.24\textwidth}
        \centering
        \includegraphics[width=\textwidth]{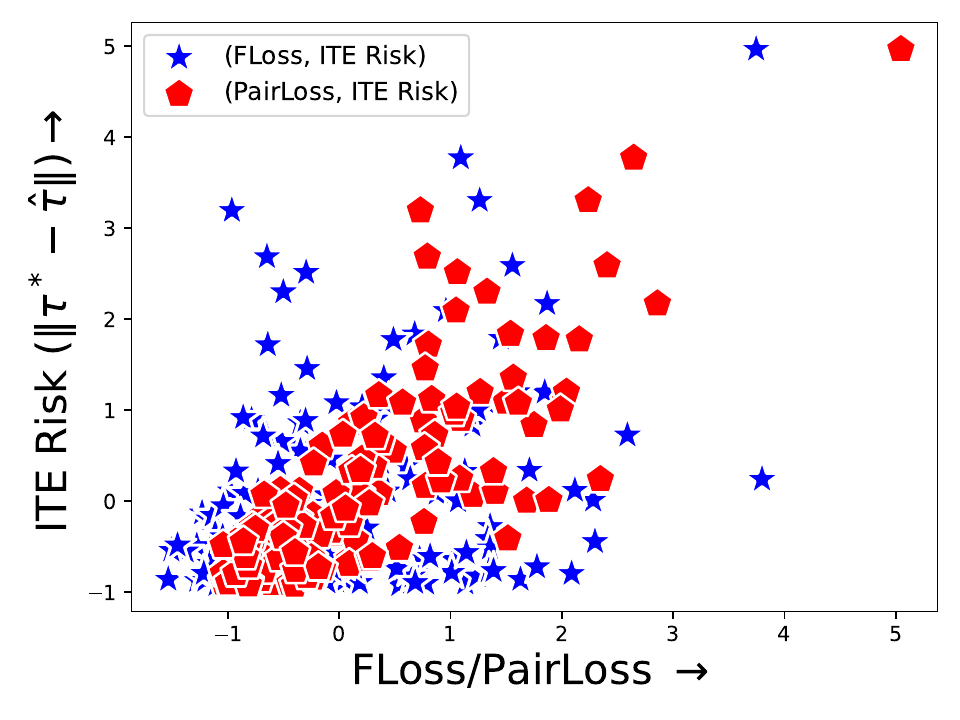} 
        \caption{Correlation of Factual \& Pair loss with the ITE risk}
        \label{fig:corr}
    \end{subfigure}
  \caption{\small{Motivating Experiment: Panel (a) presents observational data and predicted $\hat{\mu}$ functions, with pairs selected by \our\ indicated by black lines. In Panel (b), we visualize two empirical losses alongside the corresponding ITE risk. We observed a correlation of $0.32$ between factual loss and ITE risk, while \our\ achieved a substantially stronger correlation of $0.82$. Remarkably, the correlation dropped to $0.45$ when Pair loss lacked the residual alignment term, highlighting its importance.}}
  \label{fig:confound}
\end{figure}

We illustrate the working of the Pair loss in Figure~\ref{fig:mot} where we assume univariate covariates $\vx \in \RR$ (shown in black). 
Suppose we consider the fourth example $(i=4)$ that has $i=7$ in its neighborhood. The loss for this pair is computed as $((y_4-y_7)-(\hat{\mu}(x_4,0)-\hat{\mu}(x_7,1)))^2$. 
The pseudo-code for forming pairs and \our\ training is shown in Algorithm~\ref{alg:pairnet}. We defer the explanation of the algorithm to Appendix \ref{sec:algo_verbose}.
We infer the treatment effect similar to the \factual\ model using Equation~\ref{eq:floss:ite}. However, one crucial difference lies in the method of training $\hat{\mu}$.
Our loss aligns with the ITE risk defined in Eq. \ref{eq:risk}, where the only mismatch is that we impose the loss on two different covariates $\vx,\vx'$ during training. Whereas, during inference, we invoke it on the same $\vx$.  
We get further insights by expanding the Pair loss: 
\begin{align}
    \text{Pair Loss}(i, i') = & {\underbrace{
        (y_i- \hat{\mu}(\vx_i,\beta_i))^2
    }_{\text{Factual Loss}(i)}}
     + {\underbrace{(y_i'- \hat{\mu}(\vx_i',\beta_i'))^2}_{\text{Factual Loss}(i')}} \\ \nonumber
     & {\underbrace{-2(y_i- \hat{\mu}(\vx_i,\beta_i))(y_i'- \hat{\mu}(\vx_i',\beta_i'))}_{\text{Residual Alignment}(i, i')}}
\end{align}
We see that the Pair loss is decomposed into a sum of two factual losses, one on the observed samples and the other on their matched pairs, and the last term acts on error residuals. The last term promotes a positive correlation among error residuals for near covariates which is a necessary inductive bias for ITE estimation.  We will analyze the role of this term more formally in Section~\ref{sec:analyze}.

\vspace{-1mm}

\clearpage

\xhdr{A simple illustration to show alignment of Pair loss with ITE}
We present a simple study to illustrate that Pair loss is better aligned with ITE risk than \factual\ loss.  Consider binary treatments.  Let $\mu^*(\vx,0)\sim GP(0,\cK_\gamma)$ be a smooth function sampled from a Gaussian process with an RBF kernel of width $\gamma$. Next, to capture a common inductive bias~\cite{inducbias, xlearner} that the ITE function $\tau^*$ is simpler than each of the outcome functions, we model $\tau^*(\vx) \sim GP(0, \cK_\eta)$ where $\eta < \gamma$. This defines $\mu^*(\vx,1)=\mu^*(\vx,0)+\tau^*(\vx)$.  We create observation data $D$ by sampling $n_0$ instances from $\cN(u,1)$ with $t=0$, and $n_1$ instances from $\cN(u+s,1)$ with $t=1$. The value $s$ enables us to introduce confounding between the $t$ and $\vx$ values.  Figure~\ref{fig:mot} shows an example. We then sample functions from $GP(0,\cK_{\gamma'})$  and $GP(0,\cK_{\eta'})$ to serve as $\hat{\mu}(\vx,0)$ and  $\hat{\tau}(\vx)$ respectively. For each sampled function, we measure the actual ITE risk (Eq~\ref{eq:risk}), the \factual\ loss using $D$, and Pair loss using $D$. Figure~\ref{fig:corr} shows that Pair loss has a significantly higher agreement with the true ITE risk than the \factual\ loss.  
Thus, even without getting into specific network architectures and estimation methods, we observe that as a loss function, training with pairs is better suited for reducing ITE estimation errors.

\textit{\textbf{Remark}: Our proposal, Pair loss is model-agnostic, and hence we refer to any model that applies Pair loss to train for ITE estimation as \our. }

\begin{figure}[t]
    \begin{minipage}[t]{0.45\textwidth}
        \begin{algorithm}[H]
        \begin{algorithmic}[1]
            \small
            \caption{\our\ Algorithm}
            \label{alg:pairnet}
            \REQUIRE Data $D$: $\{(\vx_i, t_i, y_i)\}$, distance threshold $\delta_{\text{pair}}$, number of pairs $\text{num}_{z'}$, Epochs $E$, $\embedfct$ for forming pairs
            \STATE Let $\phi \leftarrow $ rep. network and $\{\mu_t\} \leftarrow$ prediction heads
            \STATE $D_{\text{trn}}, D_{\text{val}} \leftarrow$ \textsc{split}($D, pc=0.3$, stratify=$T$)
            \STATE $D_{\text{val}} \leftarrow$ \textsc{CreatePairDS}($D_{\text{val}}, D, \delta_{\text{pair}}, \text{num}_{z'}, \embedfct$)
            \FOR{e $\in$ [$E$]}
                \STATE $D_{\text{trn}}^e \leftarrow$ \textsc{CreatePairDS}($D_{\text{trn}}, D_{\text{trn}}, \delta_{\text{pair}}, \text{num}_{z'}, \embedfct$)
                \FOR{each batch $\{ (\vx, t, y, \vx', t', y') \} \subset D_{\text{Trn}}^e$}
                    \STATE $z,\, z' \leftarrow \phi(\vx),\, \phi(\vx')$
                    \STATE $\hat{y},\, \hat{y'} \leftarrow \mu_{t}(z),\, \mu_{t'}(z')$ 
                    \STATE $\text{loss} \leftarrow \mathcal{L}\left((y - y'),\, (\hat{y} - \hat{y'})\right)$
                    \STATE $\phi,  \{\mu_t\} \leftarrow \textsc{GradDesc}(\text{loss})$
                \ENDFOR
                \STATE Break if \textsc{EarlyStopping}($D_{\text{val}}, \phi, \{\mu_t\}$) 
            \ENDFOR
            \STATE \textbf{Return} $\phi, \{\mu_t\}$
        \end{algorithmic}
        \end{algorithm}
    \end{minipage}
    \begin{minipage}[t]{0.45\textwidth}
        \hrule
        \begin{algorithmic}[1]
        \small
            \STATE \textbf{function}  \textsc{CreatePairDS}{($D', D, \delta_{\text{pair}}$, $\text{num}_{z'}, \embedfct$)}
            \STATE $N \leftarrow |D|, D_\text{pair} \leftarrow \{\}$
            \FOR{$(\vx_i', t_i', y_i') \in D'$} 
                \STATE $d_i[j] \leftarrow \text{distance}~~ \distance(\embedfct(\vx_i'),\, \embedfct(\vx_j)) \;\; \forall j \in [N]$
                \STATE $d_i[j] \leftarrow \infty $ if $t_i' = t_j$ 
                \STATE $q_{t_i}(\vx_j|\vx_i) \leftarrow \text{softmax}(-d_i)$
                \STATE $\text{Nbrs}_i \leftarrow \textsc{sample}(q_{t_i}, \text{num}_{z'})$ 
                \STATE $D_\text{pair} \leftarrow D_\text{pair} \cup \{(x_i', t_i', y_i', x_j, t_j, y_j)\} \forall j \in \text{Nbrs}_i$
            \ENDFOR
            \STATE \textbf{Return} $D_{\text{pair}}$ after dropping largest $\delta_{\text{pair}}$ distances.
        \end{algorithmic}
        \hrule
    \end{minipage}
\end{figure}

\newcommand{\ITE}{\text{ITE}}
\newcommand{\CF}{\text{CF}}
\newcommand{\F}{\text{F}}
\newcommand{\ipm}[1]{\textsc{IPM}_{#1}}
\newcommand{\ipmg}{\ipm{G}}
\newcommand{\repspace}{\Phi}
\renewcommand{\r}{r}
\newcommand{\ferr}{\epsilon^{t}_{\F} }
\newcommand{\cferr}{\epsilon^{t}_{\CF}}
\newcommand{\pushp}[1]{p_{#1}^\phi}
\newcommand{\pairloss}{\epsilon_{\text{pair}}}
\newcommand{\pt}[1]{u_{#1}}
\newcommand{\ptreat}{\pt{1}}
\newcommand{\pcontrol}{\pt{0}}

\section{ITE Risk Bounds for Binary Treatment}
\theoremstyle{definition}
\theoremstyle{assumption}
\newtheorem*{remark}{Remark}

\label{sec:analyze}
We theoretically bound the ITE risk with \our's loss and compare it with the Factual model on binary treatments. 

\xhdr{Notation}
Recall that covariates for a treatment $t$ are sampled from $p_t(\vx) = p(\vx | t)$ and $\pt{t} = p(T = t)$ denotes treatment marginals. True outcome functions are $\mu^*_t(\xb)$ and estimated functions are $\hat{\mu}_t(\vx) = \mu(\phi(\vx), t)$.   For ease of notation, we assume that $\phi(\vx)$ is identity, but in the Appendix (Section~\ref{sec:genphi}) we show how to extend the analysis to $\phi(\vx)$ under assumptions of \cite{cfrnet}.

\begin{definition}
    The \textit{error residual} for an instance $\vx$ under a treatment $t$ is defined as $\r_t(\xb) = \hat{\mu}_t(\xb) - \mu^*_t(\xb)$.
\end{definition}
\begin{definition}
    The \textit{factual errors} under a treatment $t$ are:
    \begin{align*}
        \ferr &= \int_{\xb}{\r_t(\xb)^2 p_t(\xb) d \xb}, \;\;\;\;
        \epsilon_F = \sum_{t}\pt{t} \epsilon^{t}_{F}, 
    \end{align*}
\end{definition}

\begin{definition}
    The \textit{ITE risk} is defined in terms of residuals as:
    \begin{align*}
        \epsilon_{\ITE} 
        &= \int_{\xb}{(\r_1(\xb) - \r_0(\xb))^2 p(\xb) d \xb}
    \end{align*}
\end{definition}


\begin{definition}
For an observed instance $\xb, t$, the induced  \textit{neighborhood distribution} from which pairs $\xb'$ with $t' = 1-t$ are sampled is represented as $q_t(\xb'|\xb)$.
Marginalising over $\vx$, we obtain $q_t(\vx') = \int_{\vx}{q_t(\vx'|\vx)p_t(\vx)d\vx}$.
\end{definition}

\begin{definition}
    \our's objective in Eq.~\ref{eq:sAgree} can be written as:
    \begin{align*}
        \label{eq:pairloss_defn}
        \pairloss = \sum_{t} \pt{t} \int_{\vx} \int_{\vx'} [(\r_t(\xb) - \r_{(1 -t)}(\xb'))^2] d q_t(\vx' | \vx) d p_t(\vx)
    \end{align*}
    
\end{definition}

\newcommand{\lemmaitepair}{The difference between ITE Risk and \our\ loss can be expressed as  
    \begin{align*}
    \epsilon_{\ITE} - \pairloss = \sum_{t}\pt{(1-t)} \int_{\xb}{\r_t(\xb)^2(p_t(\xb) - q_t(\xb))d \xb}\\
    + \sum_{t} 2 \pt{t}\int_{\xb}\r_t(\xb)g_{(1-t),t}(\vx)p_t(\xb) d \xb
    \end{align*}}
\begin{lemma}
\label{theorem:lemma:ite_pair}
\lemmaitepair\
\end{lemma}

where $g_{01}(\xb) = \int_{\xb'}{(\r_0(\xb') - \r_0(\xb)) q_1(\xb' | \xb) d \xb'}$ and $g_{10}(\xb) = \int_{\xb'}{(\r_1(\xb') - \r_1(\xb)) q_0(\xb' | \xb) d \xb'}$. Here $g_{t', t}(\vx)$ denotes the expected gap between residuals of $\vx$ and its neighbours. Please refer to Appendix section \ref{app:theorem:lemma:ite_pair} for the proof.

\begin{definition}
An Integral Probability Metric (IPM) between two probability distributions $p, q$  
    given a real-valued family of functions $G$ is defined as:
    \begin{align*}
        \ipmg(p, q) = \sup_{g \in G}{\left|\int_{\xb}{g(\xb)(p(\xb) - q(\xb)) d\xb}\right|}
    \end{align*}
    Examples of IPM include Maximum Mean Discrepancy and Wasserstein distance \cite{ipm}. 
\end{definition}

\begin{assumption}[Expected distance to neighbors is bounded]
\label{as:deltanbr}
There exists a $\delta > 0$ such that $\int_{\vx'} \|\vx-\vx'\|^2q_t(\xb' | \xb) \le \delta$.
    
\end{assumption}

\newcommand{\thmitebound}{We can now bound ITE Risk with \our\ Loss as:
\begin{align*}
    \epsilon_{\ITE}  \leq \pairloss + \sum_{t} \pt{t}\left[ B\cdot\ipmg(p_t, q_t)   
    + 2K_{(1-t)}\delta \sqrt{\epsilon^{t}_{\F}}\right] 
\end{align*}
when we assume that the error residuals  $\r_0(\vx)$, $\r_1(\vx)$ are $K_0, K_1$ Lipschitz respectively,
and there exists a $B$ such that  $\frac{1}{B} \r_t(\vx)^2 \in G$,
and  the identifiability assumptions A1--A3 hold for $P(X,T,Y)$. 
}

\begin{theorem} \label{theory:theorem:ite_pair}
\thmitebound\
\end{theorem}
Please refer to Appendix section \ref{app:theory:theorem:ite_pair} for the proof.

Now, we derive conditions under which \our\ loss is a consistent estimator of ITE risk.

\newcommand{\lemmaconsistency}{Under the strict overlap assumption, \our\ is a consistent estimator of ITE.
\begin{align*}
    \lim_{N_{t} \to \infty}\epsilon_{\ITE} = 0
\end{align*}}

\begin{lemma}[Consistency of \our]
\label{theory:lemma:consistency}
\lemmaconsistency\
\end{lemma}
Please refer to Appendix section \ref{app:theory:lemma:consistency} for the proof.

Thus, for large enough $N_t$, minimising \our\ Loss drives the ITE Risk to zero. 

\subsection{Comparison with bounds of existing methods}
We compare bounds on \our's ITE risk to those of the factual model.   Theorem 1 of \cite{cfrnet} provides the bounds of the factual model as:
\begin{theorem}
   The ITE risk can be bounded with Factual loss as follows:
    \begin{align*}
      \epsilon_{\ITE}  \leq
        2(\epsilon_F^0 + \epsilon_F^1 + B \cdot  \text{IPM}_G(p_1, p_0))
    \end{align*}
when we assume that 
there exists a $B$ such that  $\frac{1}{B} \r_t(\vx)^2 \in G$,
and  the identifiability assumptions A1--A3 hold on $P(X,T,Y)$.
\end{theorem}

Comparing the bounds of the factual loss above with \our\ loss in Thm~\ref{theory:theorem:ite_pair} we make two remarks: 
\begin{remark}
First note that, 
the bound on ITE Risk with Factual loss does not go to zero even for large $N_t$ as the $\ipmg(p_0, p_1)$ would be non-zero, in general, in the presence of confounding.  In contrast, the bound with \our\ loss converges to zero under infinite samples as shown above.
\end{remark}

\begin{figure}
    \centering
    \includegraphics[width=0.4\textwidth]{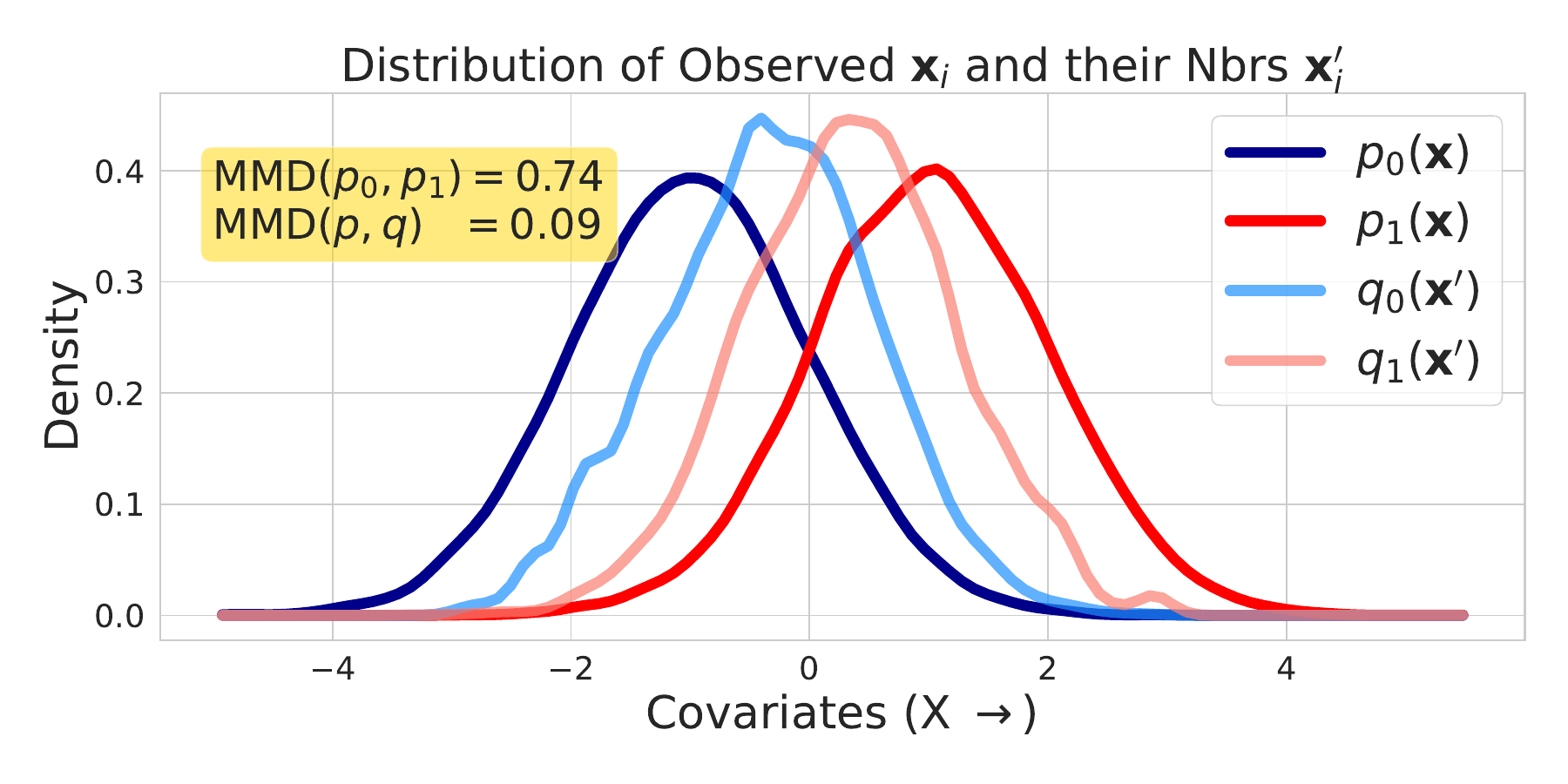}
  \caption{\small{
    We plot the distributions $p_t$ and $q_t$, where $p_0$ is $\mathcal{N}(-1, 1)$ and $p_1$ is $\mathcal{N}(+1, 1)$. We observe that the factual model relying on MMD($p_0, p_1$) shows a larger difference from the ITE risk compared to \our, which depends on MMD($p, q$).}}
  \label{fig:pq_dist}
\end{figure}

\begin{remark}
Even under finite samples, the bounds of \our\ are significantly tighter than for factual loss.  Even if we focus on the differences between the IPM terms across the two bounds, then $\ipmg(p_0,p_1)$ is observed to be larger than $\sum_t u_t\ipmg(p_t,q_t)$.  
\end{remark}
As an illustration, we consider a toy setting where the densities $p_0, p_1$ are unit variance Gaussians with means $-1, +1$ respectively.
We present the $p_t$ distribution and their induced neighbor distributions $q_t$ by \our\ in Figure \ref{fig:pq_dist}.  We observe that \our\ demonstrates a significantly lower MMD divergence of $0.09$ compared to the factual model's $0.74$. Notably, the divergence of \our\ would decrease further with increasing samples in the training dataset, while the factual model's divergence would remain the same.   Even on real datasets we observe this trend as we discuss in the experiments section.

\vspace{-1mm}

\section{Empirical Evaluation}
\label{sec:exp}

{\renewcommand{\arraystretch}{1.2}%
\begin{table*}[!h]
    \centering
    \setlength\tabcolsep{3.2pt}
    \resizebox{0.9\textwidth}{!}{
    \begin{tabular}{c|l|r|r|r|r|r|r}
    \hline
        ~ & ~ &  \multicolumn{2}{c|}{IHDP} &  \multicolumn{2}{c|}{ACIC} & \multicolumn{2}{c}{Twins} \\
        \hline
        ~ & ~ & PEHE in & PEHE out & PEHE in & PEHE out & PEHE in & PEHE out \\
        \hline \hline
        \multirow{4}{*}{\rot{25}{Meta-Learners}} & TLearner \cite{xlearner} & 1.12 (0.00) & 1.34 (0.00) & 3.54 (0.02) & 4.29 (0.03) & \first{0.32 (0.25)} & \first{0.32 (0.01)} \\
        ~ & RLearner \cite{rlearner} & 3.14 (0.00) & 3.24 (0.00) & 4.01 (0.00) & 3.94 (0.00) & \first{0.32 (0.42)} & \first{0.32 (0.15)} \\
        ~ & DRLearner \cite{drlearner} & 1.12 (0.00) & 1.35 (0.00) & \second{3.01 (0.11)} & \second{3.33 (0.08)} & \first{0.32 (0.42)} & \first{0.32 (0.14)} \\
        ~ & XLearner \cite{xlearner} & 1.77 (0.00) & 1.91 (0.00) & \second{2.90 (0.11)} & \second{3.31 (0.10)} & \first{0.32 (0.28)} & \first{0.32 (0.01)} \\
        \hline
        \multirow{3}{*}{\rot{25}{Rep. Learners}} & TARNet \cite{xlearner} & 0.74 (0.01) & \second{0.83 (0.11)} & \second{2.56 (0.26)} & \second{2.71 (0.29)} & 0.33 (0.01) & \first{0.32 (0.00)} \\
        ~ & CFRNet \cite{uri_learning_reps} & 0.96 (0.00) & 1.11 (0.00) & \second{3.12 (0.06)} & 3.45 (0.06) & 0.33 (0.00) & 0.33 (0.00) \\
        ~ & FlexTENet \cite{inducbias} & 1.03 (0.00) & 1.26 (0.00) & 3.78 (0.00) & 5.37 (0.00) & 0.37 (0.00) & 0.36 (0.00) \\
        ~ & ESCFR \cite{escfr} & 0.99 (0.12) &  1.25 (0.22)	& 2.84 (0.21)	& 3.04 (0.29) & 0.33 (0.01)	& 0.36 (0.00)\\
        ~ & StableCFR \cite{stablecfr} & 1.48 (0.00) & 1.7 (0.02) & 4.39 (0.01) & 4.53 (0.02) & 0.36 (0.00) & \first{0.32 (0.00)} \\
        \hline
        \multirow{2}{*}{\rot{25}{Weighting}} & IPW \cite{IPTW} & 0.83 (0.00) & 0.93 (0.04) & \second{2.34 (0.44)} & \second{2.57 (0.41)} & 0.33 (0.00) & 0.33 (0.00) \\
        ~ & DragonNet \cite{dragonnet} & 0.73 (0.01) & \second{0.83 (0.11)} & \second{2.57 (0.25)} & \first{2.72 (0.28)} & 0.33 (0.01) & 0.33 (0.00) \\
        \hline
        \multirow{2}{*}{\rot{25}{Matching}} & $k$NN \cite{matching_survey} & 1.34 (0.00) & 1.46 (0.00) & 3.13 (0.03) & 3.14 (0.05) & \second{0.33 (0.13)} & 0.32 (0.00) \\
        ~ & Perfect Match \cite{perfect_match} & 2.50 (0.00) & 2.66 (0.00) & 4.13 (0.00) & 4.02 (0.00) & 0.34 (0.00) & 0.33 (0.00) \\
        \hline 
        ~ & PairNet & \first{0.58 (0.00)} & \first{0.69 (0.00)} & \first{2.27 (0.00)} & \first{2.46 (0.00)} & \first{0.32 (0.00)} & \first{0.32 (0.00)} \\
        \hline
    \end{tabular}}
        \vspace{0.2cm}
       \caption{\small{RQ 1: The performance of \our\ compared to baselines evaluated using PEHE error on binary datasets. The table shows mean values and corresponding $p$-values within brackets. One-sided paired $t$-tests are conducted using \our\ as reference. Algorithms with the best mean error are highlighted in \hlc[green]{green}, while those with large $p$-values are highlighted in \hlc[yellow]{yellow}. Overall, \our\ demonstrates superior performance among all methods.}}
       \label{tab:main_bin}
\end{table*}}

We conduct experiments to address the following research questions. 
\begin{enumerate}
\item[\text{RQ1}] Does \our\ outperform state-of-the-art methods?
\item[\text{RQ2}] How sensitive is \our\ to the proximity of the pairs compared to matching methods that compute pseudo-outcomes from matched pairs?
\item[\text{RQ3}] How aligned is the Pair loss with the ITE risk (Eq. \ref{eq:risk}) on real datasets?
\item[\text{RQ4}] How sensitive is \our\ to the choice of hyper-parameters $\delta_{\text{pair}}$ and $\text{num}_{z'}$?
\item[RQ5] How does Pair loss perform when applied on other T-Learners? 
\end{enumerate}

\xhdr{Performance Metric}
We evaluate the Individual Treatment Effect (ITE) risk on a dataset $D_{\text{tst}}$, comprising counterfactual pairs represented as 5-tuples $(\xb, t, y, t', y')$. The Precision in Estimating Heterogeneous Effects (PEHE) \cite{uri_learning_reps} serves as an empirical measure of the ITE risk, defined as:
    $\sqrt{\frac{1}{|D_{\text{tst}}|}\sum\limits_{(\xb, t, y, t', y')} \left( \tau^*(\xb, t, t') - \hat{\tau}(\xb, t, t') \right)^2}$
We quantify PEHE (in) error for training instances and PEHE (out) error for test instances. Additionally, we employ hypothesis tests to evaluate the statistical significance of our results. Unlike conventional experiments that report standard deviation across runs, each seed in our experiments corresponds to a unique dataset, making hypothesis tests more appropriate, as noted in \cite{benchmarking}. We perform a one-sided paired t-test to compare \our\ performance with the baseline methods. A $p$-value below 0.05 indicates statistically significant performance improvements of \our\ over the baselines. $p$-values are enclosed in brackets in our tables.

\paragraph{Experimental Setup}
We implemented \our\ in the CATENets library \footnote{\url{https://github.com/AliciaCurth/CATENets}} using JAX. For model training, all methods reserved 30\% of the data for validation and implemented early stopping based on it. \our\ early stops on pairs as shown in Algorithm \ref{alg:pairnet}. 
We configured all hyperparameters, network architecture, optimizer, learning rate, etc. according to the CATENets defaults. The sole change was setting the weight for the $L_2$ regularizer on the $\phi$ parameters to $1$, a change applied uniformly to both \our\ and the baseline methods as it boosted the performance of all the models. Exclusive to \our\ are two hyperparameters: $\delta_{\text{pair}}$ and $\text{num}_{z'}$, set to $0.1$ and $3$, respectively, across all datasets. 
For $\embedfct$, which is used in distance computation during pair selection, we used embeddings from the representation network of a model trained until convergence on the factual loss. 

\begin{table*}[!h]
    \centering
    \setlength\tabcolsep{3.2pt}
    \resizebox{1.0\textwidth}{!}{
    \begin{tabular}{l|r|r|r|r|r|r|r|r}
    \hline
        &  \multicolumn{1}{c|}{IHDP} &  \multicolumn{1}{c|}{News} & \multicolumn{2}{c|}{TCGA-0}& \multicolumn{2}{c|}{TCGA-1}& \multicolumn{2}{c}{TCGA-2} \\
        \hline
         Training Data size  & \multicolumn{1}{c|}{$|D|$} & \multicolumn{1}{c|}{$|D|$} & \multicolumn{1}{c|}{$|D|$} & \multicolumn{1}{c|}{0.1$\times |D|$} & \multicolumn{1}{c|}{$|D|$} & \multicolumn{1}{c|}{0.1$\times |D|$} & \multicolumn{1}{c|}{$|D|$} & \multicolumn{1}{c}{0.1$\times |D|$} \\
        \hline
        \hline
        DRNet \cite{DRNet} & 2.45 (0.00) & 1.42 (0.00) & 0.34 (0.01) & 0.52 (0.00) & 0.24 (0.04) & \first{0.27 (0.53)} & \second{0.49 (0.44)} & \second{0.77 (0.06)} \\
        \our\ (DRNet) & $\first{ 2.27 (0.00)}$& $\first{1.32 (0.00)}$ & \first{0.25 (0.00)} & \first{0.44 (0.00)} & \first{0.21 (0.00)} & \first{0.27 (0.00)} & \first{0.48 (0.00)} & \first{ 0.65 (0.00)} \\
        \hline 
        VCNet \cite{vcnet}& 1.73 (0.02) &  $\first{1.24 (1.00)}$ & \first{0.25 (0.57)} & 0.43 (0.02) & \second{0.21 (0.38)} & 0.27 (0.00) & \first{0.45 (0.51)} & \second{0.58 (0.12)} \\
        \our\ (VCNet)& $\first{1.57 (0.00)}$ & 1.26 (0.00) & \first{0.25 (0.00)} & \first{0.27 (0.00)} & \first{0.21 (0.00)} & \first{0.22 (0.00)} & 0.45 (0.00) & \first{0.49 (0.00)} \\
        \hline
    \end{tabular}}
    \vspace{0.1cm}
   \caption{\small{RQ 1: Performance of Pair loss on DRNet, VCNet assessed using PEHE out error on continuous datasets. We report mean and $p$-values within brackets for a one-sided paired t-test conducted with \our\ as the baseline.}}
   \label{tab:main_cont}
\end{table*}

\subsection{RQ1: \our\ vs. Baselines}
We address this question by comparing the performance of \our\ with baselines on both binary and continuous treatments. We begin our analysis with binary treatments.
\label{sec:expt:bin}
\xhdr{Binary Datasets} We use the following three benchmark datasets: IHDP, ACIC, and Twins. The IHDP and ACIC datasets are semi-synthetic with synthetic potential outcome functions $\mu^*(\xb, t)$, while the Twins dataset contains real outcomes. We briefly describe them: 
\noindent \textbf{IHDP} The Infant Health Development Dataset 
 \cite{uri_learning_reps} contains 25 covariates, 747 examples, and 100 different realizations of the synthetic potential outcome function. 
\noindent \textbf{ACIC} The Atlantic Causal Inference Conference competition dataset (2016)\footnote{\url{https://jenniferhill7.wixsite.com/acic-2016/competition}} has 58 covariates, 4802 examples and considers 77 different potential outcome functions. Following CATENEts ~\cite{benchmarking}, we focused on three functions, namely versions 2, 7, 26, as they exhibit differences in levels of effect heterogeneity. In particular ACIC2 has no effect heterogeneity.  
\noindent \textbf{Twins} The Twins dataset \cite{cevae} provides ground truth outcomes for both treatments and has 40 covariates on 11,984 same-sex twins weighing less than $2 \text{kg}$ at birth. The treatment variable indicates which twin in each pair is heavier. The outcome variable $Y$ is binary indicating the mortality within the first year of birth.  Given the binary nature of outcomes ($Y$) in the Twins dataset, modeling the outcome difference requires a 3-way classification task, which we elaborate on in Appendix \ref{sec:app:twins}.
We defer a detailed description of the datasets to section \ref{sec:expt:bin_app} in the Appendix.

\xhdr{Binary Baselines} We group the baselines based on how they address confounding.
\textit{Meta-learners} such as TLearner \cite{xlearner}, RLearner \cite{rlearner} directly learn Individual Treatment Effects (ITE) $\tau$ after imputing pseudo-outcomes for the missing treatments.
\textit{Representation-learning} methods like TARNet \cite{Tarnet}, CFRNet \cite{cfrnet}, FlexTENet \cite{inducbias} share $\phi$ model parameters and learn treatment-specific $\mu_t$, with varied regularization.
\textit{Weighting} techniques like IPTW \cite{IPTW} impose weighted factual losses. DragonNet \cite{dragonnet} is a doubly robust method that is similar to the Augmented IPTW estimator.
\textit{Matching} approaches like $k$NN \cite{matching_survey} and Perfect Match \cite{perfect_match} perform pairing $(\vx,t,\vx',t')$ like \our\ but copy over the outcome of $\vx'$ as pseudo-outcomes for $\vx$ under $t'$ and impose loss $(\mu(\vx,t')-y')^2$. 
The pairs have to be very close for such losses not to hurt.   

\xhdr{Results on Binary Treatments} We present the results in Table \ref{tab:main_bin} and emphasize the following key observations:
\textbf{(1)} Overall, \our\ outperforms all eleven methods spanning all four categories of prior techniques for ITE. The gains on IHDP and ACIC are substantial and on Twins, all methods that model the outcome difference provide similar performance. 
\textbf{(2)} Meta Learners exhibit poor performance on IHDP and ACIC due to their two-staged regression approach, where missing outcomes are imputed in the first stage. Errors from the first stage regression are propagated to the second stage, resulting in suboptimal $\hat{\tau}$.
\textbf{(3)} Representation learners outperform meta-learners by joint training. However, these models lack the necessary inductive bias for predicting outcome differences across treatments during inference. ITE estimates can be particularly affected when error residuals for observed and alternative treatments exhibit a negative correlation. In contrast, \our\ promotes a positive correlation for nearby instances, leading to enhanced performance. 
\textbf{(4)} Weighting methods exhibit poor performance because they rely on propensity scores $\pi_t(\vx)$, which are often not well-calibrated.
\textbf{(5)} Matching methods perform poorly as they impose counterfactual losses on pseudo-outcomes, which can be unreliable.
In summary, \our\ strikes a balance by incorporating the necessary inductive biases for inferring $\tau$, while avoiding reliability issues by imposing only factual losses. Finally, we show the results for individual ACIC versions in Appendix \ref{app:ind_acic}. 
\textbf{(6)} We explain the results on the two recent baselines, StableCFR and ESCFR as follows: StableCFR results are poor because they search for pairs by partitioning the high-dimensional covariate space and conduct matching there. On the contrary, methods such as PairNet that perform pairing in low dimensional embedding $\psi$ space tend to perform better.
ESCFR relies on optimal transport for balancing the covariate representations across the treatment groups. However, OT based solutions suffer when implemented in a mini-batch on datasets with skewed treatment distributions.

\paragraph{Continuous Experiments:}
\label{sec:expt:cont}
The treatments in continuous datasets take a real value between 0 and 1. 
%
\xhdr{Continuous Datasets} \noindent \textbf{TCGA[0-2]} \cite{scigan}
dataset obtained from The Cancer Genome Atlas project consists of 4,000 covariates that represent the gene expression of 9,659 cancer patients. 
We consider three different types of cancer treatments, and the outcome variable models the risk of cancer recurrence.
\noindent\textbf{IHDP} \cite{uri_learning_reps} covariates are the same as in the discrete case but treatments and outcomes are synthetic as proposed in~\cite{vcnet}.
\noindent\textbf{News} \cite{scigan}
contains 2,858 bag-of-words covariates from 3,000 news articles taken from the New York Times. The treatment models the amount of time a user spends reading a news article while the outcomes model user satisfaction. 
 We specify the functional forms of the assumed potential outcomes and the treatment assignment mechanism in Appendix \ref{app:cont}, and a detailed discussion in Appendix \ref{sec:app:cont_datasets}. 

\xhdr{Sampling Pairs} \noindent To create pairs for continuous treatments for an observed instance $(\xb_i, t_i, y_i) \in D$, we adopt the approach mentioned in \cite{giks} to find near neighbors. We first sample $t' \in U[0, 1]$ from a uniform distribution. Then, we select pairs that are in close proximity to $\xb_i$ from within a subset of $D$ defined as $\{(\xb_j, t_j, y_j) \in D \mid |t_j - t'| < 0.05\}$; i.e., the $q_{t_i}$ distribution is defined on the filtered subset. The only change required in Algorithm \ref{alg:pairnet} for continuous treatments is in line 4, where we calculate distances $d(\xb, \xb')$ for $\xb'$ in the filtered subset.

\xhdr{Continuous Baselines}
Unlike our method, most earlier methods discussed for binary treatment do not naturally generalize to the continuous case.  For continuous treatments, two 
state-of-the-art model architectures are: DRNet \cite{DRNet} and VCNet \cite{vcnet}. Both these models share parameters in the $\phi$ layers and learn treatment-specific $\mu$ parameters. The key distinction lies in their approach to handling the treatment input. While DRNet applies binning on the $t$, VCNet employs a more sophisticated strategy of applying a spline basis expansion on $t$ and then learning of $\mu_t$ as a smooth function on it.

\xhdr{Results on Continuous Treatments}
We compare the performance of the factual loss (Eq. \ref{eq:floss}) and Pair loss using DRNet and VCNet. We present the PEHE out errors here in table \ref{tab:main_cont} and include the PEHE in errors in the Appendix section \ref{sec:app:cont_pehe_in}.
\textbf{(1)} We observe that Pair loss achieves the best mean performance across all the datasets for DRNet.
\textbf{(2)} On the VCNet model, we find that Pair loss yield a statistically significant improvement on the IHDP dataset, while factual losses perform best on the news dataset.
\textbf{(3)} For VCNet, both \our\ and factual approaches exhibit similar performance on the TCGA datasets, with some $p$-values close to $0.5$. VCNet's strong smoothness inductive bias enables factual losses to saturate performance given sufficient data (approximately 5.5k for TCGA). However, upon repeating experiments by randomly dropping about $90\%$ of the data, we observed statistically significant gains in PEHE error over the factual model. Notably, dropping data does not significantly affect \our's performance, whereas it deteriorates the Factual's performance.

\vspace{-1mm}
\subsection{RQ2: Smaller Sensitivity of \our\ to  pair proximity}
We investigate this question by changing proximity via a temperature parameter $\lambda$ to softmax at step 6 of the \texttt{CreatePairs} function in Algorithm 1; that is, $q_{t_i}(\mathbf{x}_j|\mathbf{x}_i) \leftarrow \text{softmax}(- \lambda d_i)$. Setting $\lambda = 0$ gives us random pairs, whereas increasing $\lambda$ increases the proximity of pairs.  We conduct experiments with IHDP and ACIC datasets for $\lambda$ across $\{0, 0.1, 0.25, 0.5, 1, 5\}$.  For reference,
we compare (a) a Factual model that does not involve pairing and (b) $k$NN, which imposes counterfactual losses using paired instances. 
We present the results in Figure 1. We observe that the $k$NN method is adversely affected when provided with distant pairs (small $\lambda$). In contrast, \our\ demonstrates robustness. Interestingly, even with random pairs ($\lambda=0$), \our\ performs better than the factual model because it aligns the error residuals for contrasting treatments.

We further conducted experiments with different choices for the embedding function $\embedfct$ using which the pairs are formed. We observed that \our\ performs well across a range of choices as shown in Section \ref{app:sec:phifct} in the Appendix.

\begin{figure}
    \centering
    \includegraphics[width=0.48\textwidth]{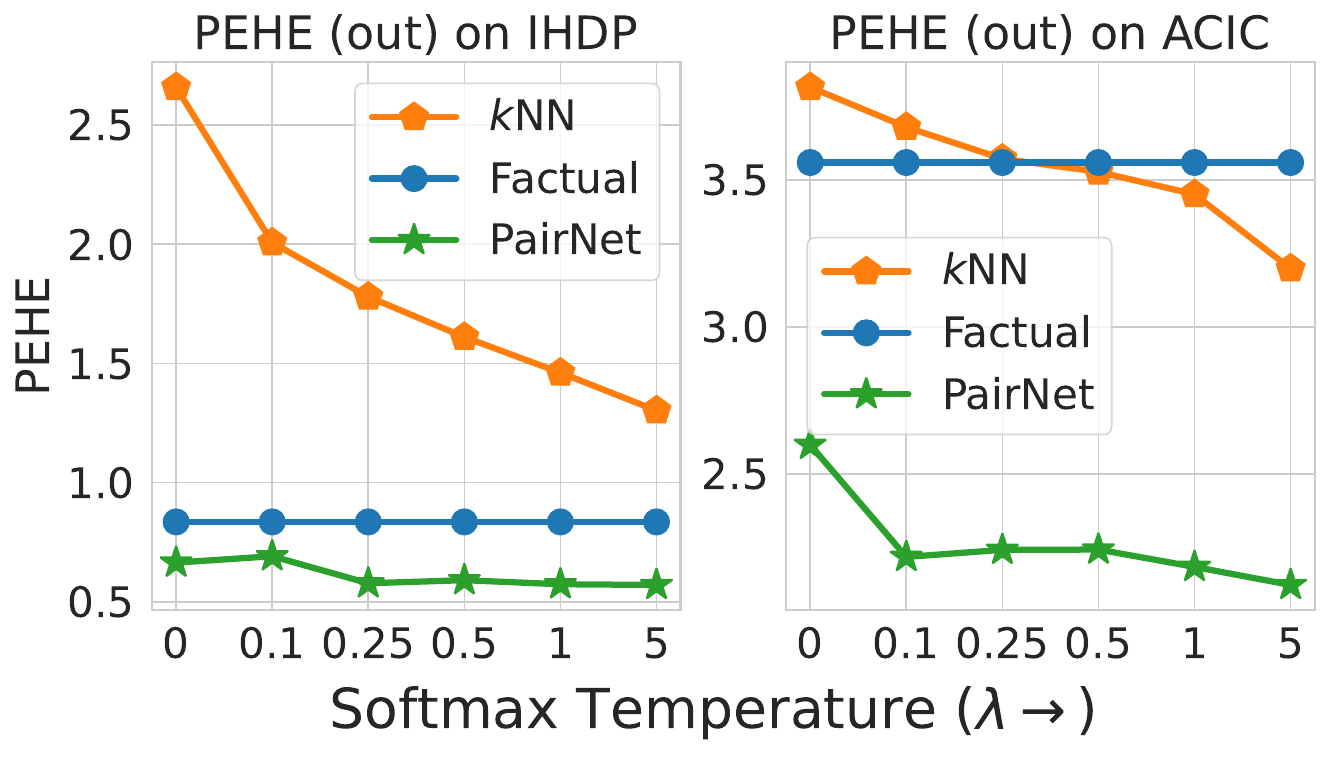}
  \caption{\small{RQ2: PEHE with increasing proximity of covariates within a pair.
  Matching methods like $k$NN deteriorate fast if pairs are not close together, whereas \our\ remains robust and provides gains over baseline also with random pairing ($\lambda=0$).}}
  \label{fig:sm_temp}
\end{figure}

\subsection{RQ3: Alignment of Pair loss with ITE Risk}
We assess the alignment on real datasets using two quantities: \textbf{(a)} correlation between Pair loss and the ITE risk, and \textbf{(b)} divergence between $p$ and $q$ distributions. These measures are computed on the test data and compared with the corresponding values achieved by the factual model. The results presented in Table \ref{tab:pq_dist} show that Pair loss exhibit a stronger correlation with the gold ITE risk.
Further, we observe that \our\ achieves smaller MMD values even on real datasets.

{\renewcommand{\arraystretch}{1.1}%
\begin{table}
    \centering
    \setlength\tabcolsep{3.2pt}
    \resizebox{.48\textwidth}{!}{
    \begin{tabular}{l|r|r|r|r}
    \hline
        \multirow{2}{*}{} &  \multicolumn{2}{c|}{Correlation of ITE Risk Eq. \ref{eq:risk}} & \multicolumn{2}{c}{Divergence} \\
        \hline
         &  Factual  Eq. \ref{eq:floss} & \our\ Eq. \ref{eq:sAgree} & MMD($p_0, p_1$) & MMD($p, q$)\\
        \hline
        \hline
        IHDP & 0.748 & \first{0.807} & 0.120 & \first{0.100} \\ 
        ACIC & 0.008 & \first{0.273} & 0.008 & \first{0.006}\\ 
        \hline
    \end{tabular}}
\vspace{0.1cm}
 \caption{\small{RQ 3: Assessing alignment between Pair loss and the ITE risk. We observe that Pair loss exhibits a stronger correlation with the ITE risk when compared with the Factual model. Additionally, \our\ consistently achieves lower MMD measures. \label{tab:pq_dist}}
 }
 \label{tab:abl:l2}
\end{table}}

\vspace{-2mm}
\subsection{RQ 4: Sensitivity of \our\ to  $\delta_{\text{pair}}$ and $\text{num}_{z'}$} \label{sec:sensitivity}
\vspace{-1mm}
The parameter $\delta_{\text{pair}}$ denotes the proportion of pairs with the largest distance excluded in \our. Meanwhile, $\text{num}_{z'}$ indicates the number of pairs selected for each training sample. We vary $\delta_{\text{pair}}$ across the range $\{0, 0.1, 0.25\}$ and $\text{num}_{z'}$ across $\{1, 2, 3, 4, 5\}$. We assess the PEHE out errors and perform a t-test using \our\ default hyperparameter as the reference method. We emphasize the default value using \hlc[cyan!50]{cyan}. The results for $\delta_{\text{pair}}$ are summarized in Table \ref{tab:abl:delta}. Interestingly, we observe that \our\ demonstrates some sensitivity to the choice of this parameter in the continuous treatment setting, although this sensitivity is less pronounced in the binary case. Remarkably, \our\ consistently outperformed the baselines across all the considered parameter values.

{\renewcommand{\arraystretch}{1.1}%
\begin{table}[]
    \centering
    \setlength\tabcolsep{3.2pt}
    \resizebox{0.45\textwidth}{!}{
    \begin{tabular}{l|r|r|r}
        \hline
         ~ &  \multicolumn{1}{c|}{0} & \multicolumn{1}{c|}{0.25} & \multicolumn{1}{c}{\third{0.1}}\\
        \hline
        \hline
        Binary & 0.48 (0.70) & 0.47 (0.62)  & 0.45 (0.00) \\ \hline
        Cont. (DRNet) & 2.26 (0.91)& 2.38 (0.61)  & 2.40 (0.00) \\
        Cont. (VCNet) & 1.41 (0.95) & 1.69 (0.36)  & 1.63 (0.00) \\
        \hline
    \end{tabular}}
    \vspace{0.1cm}
   \caption{\small{RQ 4: PEHE out error for various values of $\delta_{\text{pair}}$ on binary and continuous versions of the IHDP dataset over 5 seeds.}}
   \label{tab:abl:delta}
\end{table}}  

We present sensitivity analysis results for different choices of $\text{num}_{z'}$ in Table \ref{tab:app:numcfz}. This hyperparameter governs the number of pairs created for each observed sample, with the default value set to 3 in \our. Table \ref{tab:app:numcfz} demonstrates the robustness of \our\ to variations in $\text{num}_{z'}$ choices.

{\renewcommand{\arraystretch}{1.3}%
\begin{table}[!h]
    \centering
    \setlength\tabcolsep{3.2pt}
    \resizebox{0.5\textwidth}{!}{
    \begin{tabular}{l|r|r|r|r|r}
        \hline
         IHDP &  \multicolumn{1}{c|}{1} & \multicolumn{1}{c|}{2} & \multicolumn{1}{c|}{4} & \multicolumn{1}{c|}{5} & \multicolumn{1}{c}{\third{3}}\\
        \hline
        \hline
        Binary & 0.50 (0.48) & 0.47 (0.67) & 0.47 (0.70) & 0.44 (0.84) & 0.50 (0.00) \\
        \hline
        Cont. (DRNet) & 2.31 (0.35) & 2.29 (0.42) & 2.21 (0.69) & 2.27 (0.47) & 2.26 (0.00)\\
         
        Cont. (VCNet) & 1.46 (0.24) & 1.45 (0.28) & 1.45 (0.32) & 1.42 (0.47) & 1.41 (0.00)\\
       
        \hline
    \end{tabular}}
    \vspace{0.1cm}
   \caption{PEHE out error for various values of $\text{num}_{z'}$ on both the binary and continuous versions of the IHDP dataset. \our\ performance is similar across all the five choices.}
   \label{tab:app:numcfz}
\end{table}}

\subsection{RQ5: Pair Loss on other T-Learners}
Since Pair Loss is model-agnostic, it applies to any T-Learner-based model architecture. We chose TARNet as the default architecture due to its simplicity and lack of assumptions. In this experiment, we applied Pair Loss to other T-Learners, including CFRNet, DragonNet, and FlexTENet. The results, shown in Table \ref{tab:rq5} for IHDP and all three versions of the ACIC dataset, demonstrate that Pair Loss consistently improves performance, making it a preferable option across various model architectures.

\begin{table*}[]
    \centering
    \setlength\tabcolsep{3.2pt}
    \resizebox{0.7\textwidth}{!}{
    \begin{tabular}{l|r|r|r|r|r|r}
    \hline
        ~ & \multicolumn{2}{c|}{CFRNet} & \multicolumn{2}{c|}{DragonNet} & \multicolumn{2}{c}{FlexTENet} \\
        \hline
        Dataset & Baseline & \our & Baseline & \our & Baseline & \our \\ \hline \hline
         IHDP & 1.68 (0.33) & \first{1.03 (0.00)} & 1.23 (0.46) & \first{0.90 (0.00)} & 1.57 (0.44) & \first{1.20 (0.00)} \\
        ACIC2 & 2.12 (0.20) & \first{0.72 (0.00)} & \first{1.15 (0.34)} & 1.2 (0.00) & 4.89 (0.08) & \first{2.86 (0.00)} \\
        ACIC7 & 4.12 (0.01) & \first{3.55 (0.00)} & 3.49 (0.06) & \first{2.90 (0.00)} & 4.62 (0.12) & \first{3.85 (0.00)} \\
        ACIC26 & 4.12 (0.1) & \first{3.28 (0.00)} & 3.52 (0.3) & \first{3.32 (0.00)} & 6.60 (0.11) & \first{4.62 (0.00)} \\
        \hline
    \end{tabular}}
    \vspace{0.1cm}
   \caption{\small{RQ 5: Performance of T-Learners, including CFRNet, DragonNet, and FlexTENet, when trained with Pair loss. We compare the results of the baseline losses with our approach and observe that Pair loss provides a significant performance improvement.}}
   \label{tab:rq5}
\end{table*}

\subsection{Additional Sensitivity Analysis}

\begin{table*}[!h]
    \centering
    \setlength\tabcolsep{3.2pt}
    \resizebox{0.6\textwidth}{!}{
    \begin{tabular}{l|r|r|r|r|r}
        \hline
        $\alpha$ & 0 & 0.5 & 1 & 1.5 & \third{2} \\
        \hline
        \hline
        IHDP & \second{0.89 (0.06)} & 1.03 (0.01) & 1.03 (0.01) & 1.03 (0.01) & \first{0.69 (0.00)} \\
        ACIC2 & 0.92 (0.83) & \first{0.47 (0.97)} & 1.5 (0.53) & \second{0.51 (0.96)} & 1.56 (0.00) \\
        ACIC7 & \first{2.68 (0.68)} & 3.28 (0.19) & \second{2.89 (0.50)} & 3.25 (0.20) & \second{2.89 (0.00)} \\
        ACIC26 & \first{2.72 (0.63)} & 2.93 (0.51) & \second{2.77 (0.61)} & 2.9 (0.53) & 2.95 (0.00) \\
        \hline
    \end{tabular}}
    \vspace{0.1cm}
   \caption{\small{This table shows the results of weighting the residual term in Pair loss with $\alpha$. \our's default choice is not to weight this term (i.e., $\alpha=2$). We observe that no particular $\alpha$ achieves the best performance across all datasets with statistical significance.}}
   \label{tab:abl:residual}
\end{table*}

\noindent \textbf{Weighting the Residual Term in Pair Loss} We conducted an experiment where we adjusted the weight of the residual alignment term in Pair loss. Suppose we expand Pair loss as $(y - \hat{y})^2 + (y'-\hat{y'})^2 - \alpha (y - \hat{y})(y' - \hat{y'})$, with $\alpha=2$ as used in our paper. For Pair Loss to be non-negative we need $\alpha \in [0, 2]$. We varied $\alpha$ across $\set{0, 0.5, 1, 1.5, 2}$. The p-values represent t-tests conducted with $\alpha=2$ as the baseline.

We show the results in Table \ref{tab:abl:residual}, where we observed that no specific $\alpha$ value consistently outperformed others across the datasets. Additionally, many p-values were large, indicating that we couldn't draw conclusions on which $\alpha$ was superior with statistical significance. While we could have tuned $\alpha$ using pair error on a validation dataset, we opted to avoid hyper-parameter search and stuck to the simple squared error proposal as outlined in our paper.

\noindent \noindent \textbf{Performance across $\tau$ Complexities}
We conducted an experiment with the synthetic dataset proposed in \cite{inducbias}. The dataset comprises three variants where $\tau$ is (a)  simpler, (b) comparable, and (c) more complex than each of the $\mu_0, \mu_1$. Our results in table \ref{tab:tau_complexity} show that \our\ consistently outperforms the factual model across $\tau$ complexities.

{\renewcommand{\arraystretch}{1}
\begin{table}[]
    \centering
    \begin{tabular}{c|l|l|l}
    \hline
     $\tau$ is & (a) Simpler & (b) Comparable & (c) Complex \\
    \hline
    \hline
    Factual & 6.34 (0.00) & 7.73 (0.01) & 10.08 (0.07) \\
    \our\ & \first{5.54 (0.00)} & \first{6.22 (0.00)} & \first{08.19 ( 0.00)} \\
    \hline
\end{tabular}
\caption{\small{PEHE out errors for various complexities of the ITE function $\tau$ on synthetic data; \our\ outperforms Factual.}}
\label{tab:tau_complexity}
\end{table}}

{\renewcommand{\arraystretch}{1.1}%
\begin{table}
    \centering
    \setlength\tabcolsep{3.2pt}
    \resizebox{0.48\textwidth}{!}{
    \begin{tabular}{l|r|r|r|r}
    \hline
        ~ &  \multicolumn{2}{c|}{TARNet} & \multicolumn{2}{c}{\our} \\
        \hline
         ~ &  \multicolumn{1}{c|}{1e-4} & \multicolumn{1}{c|}{\third{1}} & \multicolumn{1}{c|}{1e-4} & \multicolumn{1}{c}{\third{1}}\\
        \hline
        \hline
        IHDP & 1.51 (1.00) & \first{0.83 (0.00)} & 0.74 (0.35) & \first{0.69 (0.00)} \\ 
        ACIC(2) & 3.05 (0.74) & \first{2.71 (0.00)} & 3.52 (1.00) & \first{2.46 (0.00)} \\
        \hline
    \end{tabular}}
\vspace{0.1cm}
 \caption{\small{PEHE out error for various values of $L_2$ penalty applied on the $\phi$ parameters. A weight of 1 achieves better results.
 }}
 \label{tab:abl:l2}
\end{table}}

\noindent \textbf{Impact of $L_2$ Penalty}
In Table \ref{tab:abl:l2}, we illustrate the significance of applying an $L_2$ penalty to the $\phi$ parameters. We explore two different magnitudes of imposing the $L_2$ loss: 1e-4, the default value for CATENets, and a value of $1$. We observed consistent enhancements across all the baselines, including \our by introducing a large penalty to the $\phi$ parameters. This improvement was particularly noticeable for \our\ with version 2 of the ACIC dataset. For a detailed analysis of this hyperparameter, please refer to Appendix Section \ref{sec:app:l2}.

\section{Conclusion}  
In this paper, we introduced \our, a simple yet effective approach for estimating treatment effects. \our\ is model-agnostic, applicable to diverse treatment domains, and can be integrated with most prior networks. Our key idea was to only impose factual losses on pairs of neighboring instances. We showed that this approach effectively aligns error residuals for the chosen pairs, thereby aiding in ITE inference when outcomes are predicted for the same instance under two treatments. Across various benchmarks involving both discrete and continuous treatments, \our\ showcased significant improvements over most existing methods.
Furthermore, we theoretically characterized the difference between Pair loss and the ITE risk, showing that this difference depends on the proximity between instances in the selected pairs. This implies that under certain overlap assumptions, given a large training dataset, Pair loss serves as a consistent estimate of the ITE risk. Additionally, through several experiments and sensitivity analyses, we highlighted the merits of \our\ and provided insight into the superior performance achieved by \our\ over the baselines.

\clearpage

\section*{Acknowledgement}

Lokesh acknowledges funding from the Prime Minister Research Fellowship, Government of India, and Microsoft Research India for the travel award.

\section*{Impact Statement}
Developing new and effective methods like \our\ to understand how treatments impact different situations is crucial for decision-making across various fields like healthcare, economics, and education. For instance, in healthcare, precise knowledge about how treatments affect individuals can lead to better, more personalized medical care, ultimately improving access to quality healthcare. Similarly, in economics, understanding the consequences of policies can help policymakers make informed decisions that benefit society.

However, while these methods provide valuable insights, it is crucial to address potential ethical concerns, particularly regarding fairness. Before deploying these methods, it is essential to ensure that the observational dataset used is unbiased and does not unfairly benefit certain groups of individuals, which could lead to biased effect prediction during inference. Addressing such issues is essential for the responsible deployment of these methods and their positive impact on decision-making and societal outcomes.

\bibliography{refs}
\bibliographystyle{icml2024}

\newpage
\appendix
\onecolumn
\appendix
\onecolumn

\begin{center}
    \Large { \bf Appendix \\
    \normalsize (PairNet: Training with Observed Pairs to Estimate Individual Treatment Effect)
    }
\end{center}

\section{Explanation for \our\ Algorithm}
\label{sec:algo_verbose}
We explain the pseudocode in Alg. \ref{alg:pairnet} as follows:

\begin{itemize}
    \item  At line 2, the training dataset is split into $D_{\text{trn}}$ and $D_{\text{val}}$, stratified by treatments. For continuous $T$, a random split is used.

    \item  At line 3, pairs are created to compute validation performance for early stopping. It is important to note that pairs for samples in the validation dataset can also be obtained from $D_{\text{trn}}$. This approach is taken because $D_{\text{val}}$ is sparse, and searching for pairs within it may result in many distant pairs, which would not accurately reflect true validation performance.

    \item During each epoch $e$, pairs ($D_{trn}^e$) for imposing pair loss are created (line 5) using the CreatePairDS procedure. For training, pairs are created within the $D_{\text{trn}}$ split.

    \item The CreatePairDS procedure samples $num_{z’}$ neighbors based on the $q_t$ distribution, defined as a softmax over negative distances (line 6).

    \item Line 10 includes $\delta_{\text{pair}}$ to avoid imposing losses on extremely distant pairs. Training continues using standard mini-batch gradient descent on Pair loss (lines 7-10).

    \item  Once paired samples for training are obtained, the pair loss is imposed, and gradient descent is performed until convergence, monitored by the pair loss on $D_{\text{val}}$.
\end{itemize}

\xhdr{Intuition:} The main intuition behind PairNet is to better correlate with ITE loss by fitting the difference in outcomes of nearby covariates. This is further explained using Equation \ref{eq:sAgree} and Figure 1. We argue that Pair Loss aligns better with ITE loss because its ITE generalization gap (Theorem 5.9) is bounded by $D(p_t(X) || q_t(X))$, where $p_t(X)$ represents observed covariates for $T=t$ and $q_t(X)$ represents paired neighbor distribution. This is much lower than that of a factual model, bounded by $D(p_0||p_1)$, as depicted in Figure 2 and Thm 5.12.

\section{Theoretical Analysis}

\begin{remark}
\label{sec:genphi}
    For ease of notation, in the main paper, we showed results when the representation extraction $\phi$ is an identity map. In the Appendix, we present a generalized version of our results under certain assumptions on $\phi$ which we elaborate on now. Note that this modification does not affect the form of bounds that we obtain for Factual or \our.
\end{remark}

\begin{assumption}[$\phi$ is diffeomorphic] \label{theory:assm:phi}
    The embedding function $\phi$ is a twice-differentiable, invertible function, serving as a push-forward operator between the spaces $\cX$ and $\repspace$.
\end{assumption}

\begin{definition}[Covariate distribution under $\phi$]
    Under the push-forward operator $\phi$, for an instance $e = \phi(\vx)$, we define $\pushp{t}(e) = p_t(\phi^{-1}(e)) \left|\frac{d \phi^{-1}(e)}{de}\right|$. 
\end{definition}

\begin{assumption}[Instance error lies in $G$]
\label{theory:assm:G}
    Let $G$ denote a family of functions $\{g : \repspace \mapsto \RR\}$. There exists a constant $B_\phi$ such that for any $e = \phi(\vx), t$, we have $\frac{1}{B_\phi} \r_t(\phi^{-1}(e))^2 \in G$, where $\r_t(\vx)^2$ is the squared error. 
\end{assumption}

In the following proofs, $\ipmg$ is modified to consider a function space $G$ over $\repspace$ instead of $\cX$ in the main paper. Thus all bounds with $B\cdot\ipmg(p_t, q_t)$ change to $B_{\phi}\cdot\ipmg(p^{\phi}_t, q^{\phi}_t)$

\begin{remark}
    The original bounds in \cite{cfrnet} features a $\sigma_Y^2$ term that stems from the irreducible noise in the observed outcomes $y_i$. However, we drop that term in our analysis for brevity. 
\end{remark}

\subsection{Proof of Lemma \ref{theorem:lemma:ite_pair}} \label{app:theorem:lemma:ite_pair}

\lemmaitepair\
\xhdr{Proof}

\begin{align*}
    \epsilon_{\ITE} &= \int_{\xb}{(\r_1(\xb) - \r_0(\xb))^2 p(\xb) d \xb}\\
    &= \int_{\xb}{(\r_1(\xb) - \r_0(\xb))^2 (p_0(\xb)\pcontrol + p_1(\xb)\ptreat) d \xb} = \pcontrol\int_{\xb}{(\r_1(\xb) - \r_0(\xb))^2 p_0(\xb)d \xb}
    + \ptreat\int_{\xb}{(\r_1(\xb) - \r_0(\xb))^2 p_1(\xb)d \xb}\\
    &= \pcontrol\int_{\xb}{(\r_1(\xb)^2 + \r_0(\xb)^2 - 2\r_1(\xb)\r_0(\xb)) p_0(\xb)d \xb}
    + \ptreat\int_{\xb}{(\r_1(\xb)^2 + \r_0(\xb)^2 - 2\r_1(\xb)\r_0(\xb)) p_1(\xb)d \xb} \\
    \pairloss &= \int_{\xb} \int_{\xb'}{(\r_1(\xb) - \r_0(\xb'))^2 q_1(\xb' | \xb)p_1(\xb) \ptreat  d\xb d\xb'}
    + \int_{\xb} \int_{\xb'}{(\r_1(\xb') - \r_0(\xb))^2 q_0(\xb' | \xb)p_0(\xb) \pcontrol  d\xb d\xb'}\\
    &= \int_{\xb} \int_{\xb'}{(\r_1(\xb)^2 + \r_0(\xb')^2 - 2\r_1(\xb)\r_0(\xb')) q_1(\xb' | \xb)p_1(\xb) \ptreat  d\xb d\xb'} \\
    &\!\!+ \int_{\xb} \int_{\xb'}{(\r_1(\xb')^2 + \r_0(\xb)^2 - 2\r_1(\xb')\r_0(\xb)) q_0(\xb' | \xb)p_0(\xb) \pcontrol  d\xb d\xb'} \\
    &= \int_{\xb} {\r_1(\xb)^2 p_1(\xb) \ptreat  d\xb + \int_{\xb'} {\r_0(\xb')^2 q_1(\xb') \ptreat d \xb'} - \int_{\xb} \int_{\xb'} 2\r_1(\xb)\r_0(\xb') q_1(\xb' | \xb)p_1(\xb) \ptreat  d\xb d\xb'} \;\;\;\; \\ 
    & \quad \text{(By marginalizing over variables)}\\
    &+ \int_{\xb} \r_0(\xb)^2 p_0(\xb) \pcontrol  d\xb + \int_{\xb'} {\r_1(\xb')^2 q_0(\xb') \pcontrol d \xb'} - \int_{\xb} \int_{\xb'} {2\r_1(\xb')\r_0(\xb) q_0(\xb' | \xb)p_0(\xb)\pcontrol  d\xb d\xb'}
\end{align*}

Now taking the difference between both expressions and cancelling common terms, we obtain the lemma. 

\subsection{Proof of Theorem \ref{theory:theorem:ite_pair}}
\label{app:theory:theorem:ite_pair}
\thmitebound\

\xhdr{Proof}
From Lemma \ref{theorem:lemma:ite_pair}, we have
    \begin{align*}
    \epsilon_{\ITE} - \pairloss &= \pcontrol\int_{\xb}{\r_1(\xb)^2(p_0(\xb) - q_0(\xb))d \xb} + \ptreat \int_{\xb}{\r_0(\xb)^2(p_1(\xb) - q_1(\xb))d \xb} \\
    &+ 2\ptreat\int_{\xb}\r_1(\xb)g_{01}(\vx)p_1(\xb) d \xb + 2\pcontrol\int_{\xb}\r_0(\xb)g_{10}(\vx)p_0(\xb) d \xb  
    \end{align*}

    \begin{align*}
    \epsilon_{\ITE}  &\leq \pairloss + {\underbrace{\pcontrol\left|\int_{\xb}{\r_1(\xb)^2(p_0(\xb) - q_0(\xb))d \xb}\right| + \ptreat \left|\int_{\xb}{\r_0(\xb)^2(p_1(\xb) - q_1(\xb))d \xb}\right|}_{\text{Term 1}}} \\
    &+ {\underbrace{2\ptreat \left|\int_{\xb}\r_1(\xb)g_{01}(\vx)p_1(\xb) d \xb\right| + 2\pcontrol\left|\int_{\xb}\r_0(\xb)g_{10}(\vx)p_0(\xb) d \xb\right|}_{\text{Term 2}}}
    \end{align*}

Using $\frac{1}{B_\phi} \r_t(\vx)^2 \in G$, we can bound Term 1 as follows:

\begin{align*}
    \left|\int_{\xb}{\r_1(\xb)^2(p_0(\xb) - q_0(\xb))d \xb}\right| &= \left|B_{\phi}\int_{\Phi}{\frac{1}{B_{\phi}}\r_1(\phi^{-1}(e))^2(p^{\phi}_0(e) - q^{\phi}_0(e)d e}\right| \\
    &\leq B_{\phi}\sup_{g \in G}\left|\int_{\Phi}(g(e)(p^{\phi}_0(e) - q^{\phi}_0(e))de\right| = B_{\phi}\ipmg(p^{\phi}_0, q^{\phi}_0)
\end{align*}

Thus, Term 1 can be upper-bounded by $\pcontrol B_{\phi}\ipmg(p^{\phi}_0, q^{\phi}_0) + \ptreat B_{\phi} \ipmg(p^{\phi}_1, q^{\phi}_1)$.

Recall that $g_{01}(\xb) = \int_{\xb'}{(\r_0(\xb') - \r_0(\xb)) q_1(\xb' | \xb) d \xb'}$ and $g_{10}(\xb) = \int_{\xb'}{(\r_1(\xb') - \r_1(\xb)) q_0(\xb' | \xb) d \xb'}$.

We now derive an upper bound for Term 2 as follows:

\begin{align*}
    \left|\int_{\xb}\r_1(\xb)g_{01}(\vx)p_1(\xb) d \xb\right|^2 \leq & \int_{\xb}\r^2_1(\xb)p_1(\xb)d \xb \int_{\xb}(g_{01}(\xb))^2p_1(\xb)d \xb  \;\;\;\; \text{(By Cauchy-Schwarz Inequality)} \\
    \leq & \int_{\xb}\r^2_1(\xb)p_1(\xb)d \xb \int_{\xb}(K^2_0 \delta^2) p_1(\xb)d \xb \;\;\;\; \\
    = & \epsilon^{1}_{\F} K^2_0 \delta^2
\end{align*}

Here we use the fact that $g_{t', t}(\xb)$ can be bounded as follows under the Lipschitz continuity assumption
\begin{align*}
    |g_{01}(\vx)| \leq K_0 \delta ~~~~     |g_{10}(\vx)| \leq K_1 \delta
\end{align*}

\begin{align*}
    |g_{01}(\xb)| = \left|\int_{\xb'}{(\r_0(\xb') - \r_0(\xb)) q_1(\xb' | \xb) d \xb'}\right| \leq \int_{\xb'}{\left|(\r_0(\xb') - \r_0(\xb))\right| q_1(\xb' | \xb) d \xb'}
\end{align*}

Using Lipschitz continuity $\left|(\r_0(\xb') - \r_0(\xb))\right| \leq K_0 \Vert \xb' - \xb \Vert$, $\left|(\r_1(\xb') - \r_1(\xb))\right| \leq K_1 \Vert \xb' - \xb \Vert$

\begin{align*}
    |g_{01}(\xb)| \leq \int_{\xb'}{K_0 \Vert \xb' - \xb \Vert q_1(\xb' | \xb) d \xb'}
\end{align*}

Now using \ref{as:deltanbr}, the expected neighbour distance is bounded by $\delta$ yielding $|g_{01}(\xb)| \leq K_0 \delta$. By symmetric arguments we can bound $|g_{10}(\xb)| \leq K_1 \delta$

Thus, $\left|\int_{\xb}\r_1(\xb)g_{01}(\vx)p_1(\xb) d \xb\right| \leq  K_0 \delta \sqrt{\epsilon^{1}_{\F}}$. We can now bound Term 2 with $2\ptreat K_0 \delta \sqrt{\epsilon^{1}_{\F}} + 2\pcontrol K_1 \delta \sqrt{\epsilon^{0}_{\F}}$



\subsection{Proof of Lemma \ref{theory:lemma:consistency}}
\label{app:theory:lemma:consistency}

 \lemmaconsistency\
 
\xhdr{Proof}
We prove the consistency of \our\ under strict overlap \cite{d2021overlap}. First, we show that strict overlap can be used to derive a lower bound on the radius of a sphere around any point in which one can find a sample with the opposite treatment with high probability. Next, we show that this lower bound decreases with number of samples, $m$, allowing us to shrink the radius in which we find neighbours, thereby ensuring $q_t$ distribution converges to $p_t$.

Strict overlap states that $c < p(t|x) < 1 - c$ for some $c > 0$. Consider any sample $x$ with treatment $t$. For pairing, we need to find a matching sample $x'$ with alternative treatment $t'=1-t$ from the $m$ observations in the dataset that are assigned treatment $t'$.

Define $\bar{p}_r = \int_{\xb' \in \mathfrak{B}_r(\xb)}{p(\xb'|t’) d\xb'}$ as the probability mass of $p(X = \xb’|T = t’)$ in a ball of radius $r$ around $x$ where $\mathfrak{B}_r(\boldsymbol{\xb}) := \set{\xb' | \xb' \in \mathbb{R}^d, ||\xb' - \xb|| < r}$.

The probability that at least one sample $\xb'_j$ lies within this ball is $\bar{p}(r, m) = 1 - (1 - \bar{p}_r)^m$. If $\lim_{m \to \infty} \bar{p}(r, m) = 1$, then \our\ would find a near neighbor within this ball. This is satisfied for $\bar{p}_r = \omega(m^{-1})$ (where $\omega$ denotes asymptotic lower bound).

Now, we construct a condition for $\bar{p}_r = \omega(m^{-1})$. First note that $\bar{p}_r > \text{Volume}(\mathfrak{B}_r(x)) \cdot \min_x' p(x'|t')$ (where the minimum is over the $x' \in \mathfrak{B}_r(x)$) and $\text{Volume}(\mathfrak{B}_r(x)) \propto r^d = kr^d$.

$\min_{x' \in \mathfrak{B}_r(x)}p(x'|t') = \min_{x' \in \mathfrak{B}_r(x)} \frac{p(t’|x')p(x')}{p(t’)} > \frac{c}{p(t’)} \min_{x' \in \mathfrak{B}_r(x)}p(x')$ (By Strict Overlap)

Combining these two, we get $\bar{p}_r > \frac{k c}{p(t’)}r^d\min_{x' \in \mathfrak{B}_r(x)}p(x')$

If we shrink the ball around $x$ as $r = \omega(m^{-1/d}), \lim_{m \to \infty}r = 0$ then we get the desired condition $\bar{p}_r = \omega(m^{-1})$. 

Therefore, for any $\xb, t$ we can sample a neighbour $\xb', t' = 1-t$ within $\mathfrak{B}_r(x)$ for $r = \omega(m^{-1/d})$ with high probability $\bar{p}(r, m)$.

Since we set $r = \omega(m^{-1/d})$ which decreases with $m$, we have $\lim_{m \to \infty}r = 0$ and $\lim_{m \to \infty}\bar{p}(r, m) = 1$. Thus, $q_t(\vx' | \vx)$ converges to a dirac-delta distribution $\delta_0(\vx' - \vx)$. Thus, the marginal $q_t(\vx') = p_t(\vx')$ and $\ipmg(p_t, q_t) = 0$. Also, the $\delta$ in Assumption~\ref{as:deltanbr} converges to 0.

\section{Detailed Description of the Datasets}
Here, we present a detailed description of the various datasets used in our work.
\subsection{Binary Treatments}
\label{sec:expt:bin_app}

\begin{table}[h]
\centering 
\setlength{\extrarowheight}{2pt}
\begin{tabularx}{\linewidth}{c|r|r|r|>{\RaggedRight\arraybackslash}X|>{\raggedright\arraybackslash}X|r|>{\raggedright\arraybackslash}X}
\hline
\textbf{Dataset} & \textbf{Covariates} & \textbf{Samples} & \textbf{Runs} & \textbf{Covariates Type} & \textbf{Treatment} &  \textbf{Synthetic $\mathcal{Y}$}? & \textbf{Outcome}\\
\hline \hline
IHDP  & 25 & 747 & 100 & Features of an infant & Intervention by specialist doctor & \cmark & Cognitive test scores \\
\hline
ACIC  & 58 & 4802 & 30 &  Demographic, clinical features, etc. of pregnant women & N/A & \cmark & Developmental disorders \\
\hline
Twins  & 40 & 11,984 & 100 & Characteristics of same-sex twins both weighing less than $2$kg & Heavier twin & \xmark & One-year infant mortality\\
\hline
\end{tabularx}
\caption{This table provides information about the Binary datasets used in our work. All these datasets exhibit treatment selection bias, meaning that the observed treatments are influenced by the covariates. The 'Runs' column indicates the number of experiments conducted for each dataset, from which the associated $p$-values are calculated. Notably, the ACIC dataset encompasses three distinct potential outcome functions (versions 2, 7, 26), each repeated for 10 seeds, resulting in a total of 30 runs. For the cases where $\mathcal{Y}$ is synthetic, the assumed potential outcome functions that generate the observed outcomes can at most be interpreted as modeling the original intended outcomes.}
\end{table}

\xhdr{IHDP} 
The Infant Health and Development Program (IHDP) is a randomized controlled trial designed to evaluate the impact of physician home visits on the cognitive test performance of premature infants. The dataset exhibits selection bias, as non-random subsets of treated individuals are deliberately removed from the training dataset. Since we have observed outcomes for only one treatment, to render the dataset suitable for causal inference, we generate both observed and counterfactual outcomes using a synthetic outcome generation function based on the original covariates considering both treatments. 
The IHDP dataset comprises 747 subjects and includes 25 variables. While the original dataset discussed in \cite{cfrnet} had 1000 versions, a smaller version of the dataset with 100 versions is used in our work, aligning with the CATENets benchmark. 
Each version varies in terms of the complexity of the assumed outcome generation function, treatment effect heterogeneity, etc. As outlined in \cite{benchmarking}, reporting the standard deviation of performance across the 100 different seeds is inappropriate and therefore we calculate $p$-values through paired t-tests between our method (\our) and other baseline methods such that \our\ serves as the baseline for all experiments. Specifically, we accept the hypothesis that \our\ is superior to the baseline if the resulting $p$-value is less than 0.05.

\xhdr{ACIC}
The Atlantic Causal  The Atlantic Causal Inference Conference competition dataset (2016)\footnote{\url{https://jenniferhill7.wixsite.com/acic-2016/competition}} contains a total of 77 datasets. The covariates in all these datasets are the same and contain 58 features obtained from a real study called the Collaborative Perinatal Project. 
Each dataset involves simulating binary treatment assignments and continuous outcome variables. The datasets exhibit variations in several aspects, including the complexity of the treatment assignment mechanism, treatment effect heterogeneity, the ratio of treated to control observations, overlap between treatment and control groups, the dimensionality of confounder space, and the magnitude of the treatment effect. 
All datasets share common characteristics such as independent and identically distributed observations conditional on covariates, adherence to the ignorability assumption (selection on observables with all confounders measured and no hidden bias), and the presence of non-true confounding covariates. Of these 77 datasets, we opted to work with a subset of three datasets, specifically versions 2, 7, and 26, aligning with the CATENets benchmark. These three settings present non-linear covariate-to-outcome relationships and showcase maximum variability in terms of treatment effect heterogeneity. 
Notably, version 2 exhibits no heterogeneity, i.e. the treatment effect remains constant across all individuals. However, accurately estimating the outcome differences even for this version proves challenging as algorithms find it difficult to overcome the inherent noise observed in potential outcome realizations in the dataset. Specifically, in \our\, we found that coming up with good pairs is important as we will explain in detail in the Table \ref{tab:abl:phifct} in our main paper. The other two dataset versions show medium and high heterogeneity in terms of the treatment effects across the individuals.

\xhdr{Twins} The Twins dataset \cite{cevae} stands out as the sole dataset with actual observed outcomes. This study operates on the premise that the two twins within each pair share equivalence across all covariates, differing solely in terms of their treatment assignments. 
This unique characteristic allows the dataset to be employed as is for causal inference tasks. The dataset encompasses a total of 11,984 pairs of twins and focuses on one-year mortality as a function of birth weight, serving as the underlying treatment variable. 
To ensure the covariate equivalence, the study exclusively includes same-sex twins with birth weights below $2 \text{kg}$. In total, the dataset incorporates 39 relevant covariates. The dataset's outcomes are binary and exhibit a class imbalance in the observed outcomes. Thankfully, the mortality rates are low and stand at 16.1\% for the treated group and 17.7\% for the untreated group. Consequently, observing twin pairs with opposite outcomes in the dataset is a rare occurrence. 
In our experiments, we allocate 50\% of the dataset for testing purposes. To introduce imbalance in the treated vs. control examples in the training dataset, we sample the treated group for each twin pair using probabilities from the set $\{0.1, 0.25, 0.5, 0.75, 0.9\}$. 
For each of these probabilities, we explore the sample efficiency of various methods by varying the number of training examples across the range of $\{500, 1000, 4000, 5700\}$. These experiment settings closely match that of the CATENets benchmark.

\subsection{Continuous Treatments} \label{sec:app:cont_datasets}

In this section, we provide a more detailed description of the five continuous benchmark datasets used in our work.

\begin{table}[h]
\centering
\begin{tabular}{p{0.09\linewidth}|p{0.1\linewidth}|p{0.07\linewidth}|p{0.05\linewidth}|p{0.18\linewidth}|p{0.18\linewidth}|p{0.15\linewidth}}
\hline
\textbf{Dataset} & \textbf{Covariates} & \textbf{Samples} & \textbf{Runs} & \textbf{Covariates Type} & \textbf{Dosage} & \textbf{Outcome}\\
\hline \hline
TCGA (0--2) \cite{scigan} & 4000 & 9659 & 10 & Gene expressions of cancer patients  & dosage of the drug & cancer recurrence \\
\hline
IHDP \cite{vcnet} & 25 & 747 & 50 & features of an infant & Amount of intervention by a specialist doctor &  cognitive test scores \\
\hline
News \cite{news} & 2858 & 3000 & 20 & Bag-of-words from news articles & time spent reading the news article & user-satisfaction\\
\hline
\end{tabular}
\caption{This table shows the information on the Continuous datasets where the treatment assignment and the corresponding responses are synthesized. The assumed potential outcome functions that generate the observed $y$ can at most be interpreted as modeling the original intended outcomes.}
\end{table}

\xhdr{TCGA (0--2) \cite{scigan}}
The TCGA dataset, sourced from The Cancer Genome Atlas project, contains information on various cancer types for a total of $9659$ individuals. Each individual is characterized by $4000$ dimensions of gene expression covariates. These covariates have been log-normalized and subsequently normalized to achieve unit variance. The treatment variable signifies the dosage of the drug administered to the patient, while the synthetic response models the risk of cancer recurrence. In our experiments, we make use of three versions of the TCGA dataset introduced in \cite{scigan}, denoted as TCGA(0), TCGA(1), and TCGA(2).

\xhdr{IHDP}
The IHDP dataset was originally collected as part of the Infant Health Development Program for binary treatment effect estimation. Treatments in this dataset were assigned through a randomized experiment and it comprises $747$ subjects with $25$ covariates. For the continuous treatment effect problem, the dataset was adapted in \cite{vcnet} by assigning synthetic treatments and targets.

\xhdr{News \cite{news}}
The News dataset was initially designed for binary treatment effect estimation and was adapted in \cite{scigan} to accommodate continuous treatments and targets. In this dataset, the treatment variable represents the time spent by a user reading a news article, while the synthetic response aims to mimic user satisfaction. The dataset comprises $3000$ randomly selected articles from the New York Times, with $2858$ bag-of-words covariates.

For each of these datasets, we generate multiple versions using different random seeds, as per previous research.

\subsection{Continuous Dosage and Response Generation Functions} \label{app:cont}
We adopt the dataset generation procedures from previous methods and \textbf{quote} a detailed explanation of the process here for completeness, following \cite{DRNet, vcnet, scigan}.

\xhdr{IHDP}
For treatments in the interval $[0,1]$, we generate responses as follows:
\begin{align}
    \tilde{t} \given \xb &= \frac{2x_1}{1+x_2} + \frac{2\max(x_3,x_5,x_6)}{0.2+\min(x_3,x_5,x_6)} + 2\tanh\left(\frac{5\sum_{i\in S_{\text{dis},2}}(x_i-c_2)}{|S_{\text{dis},2}|}-4+\mathcal{N}(0,0.25)\right), \\
    t &= (1+\exp(-\tilde{t}))^{-1}, \\
    y \given \xb, t & = \frac{\sin(3\pi t)}{1.2-t}\cdot\tanh\left(\frac{5\sum_{i\in S_{\text{dis},1}}(x_i-c_1)}{|S_{\text{dis},1}|}\right) + \frac{\exp(0.2(x_1-x_6))}{0.5+5\min(x_2,x_3,x_5)} + \mathcal{N}(0,0.25),
\end{align}
where $S_{\text{con}}=\{1,2,3,5,6\}$ is the index set of continuous covariates, $S_{\text{dis},1}=\{4,7,8,9,10,11,12,13,14,15\}$, $S_{\text{dis},2}=\{16,17,18,19,20,21,22,23,24,25\}$, and $S_{\text{dis},1}\bigcup S_{\text{dis},2}=[25]-S_{\text{con}}$. Further, $c_1=\mathbb{E}\left[\frac{\sum_{i\in S_{\text{dis},1}} x_i}{|S_{\text{dis},1}|}\right]$ and $c_2=\mathbb{E}\left[\frac{\sum_{i\in S_{\text{dis},2}} x_i}{|S_{\text{dis},2}|}\right]$.

\xhdr{News}
For the News dataset, we first generate $v_1', v_2', v_3'$ from $\mathcal{N}(0,1)$ and set $v_i= \frac{v_i'}{\| v_i' \|}$. The treatment and response generation is as follows:
\begin{align}
    t \given \xb &= \text{Beta}\left(2, \left\| \frac{v_3^{\top}}{2v_2^{\top}\xb} \right\|\right), \\
    y' \given \xb, t &= \exp\left(\frac{v_2^{\top}\xb}{v_3^{\top}\xb}-0.3\right), \\
    y \given \xb, t &= 2(\max(-2,\min(2,y'))+ 20v_1^{\top}\xb)  (4(t-0.5)^2 + \sin\left(\frac{\pi}{2}t\right) + \mathcal{N}(0,0.5) \label{eq:news_response}.
\end{align}

\xhdr{TCGA(0-2)}
For TCGA(0-2), we first generate $v_1', v_2', v_3'$ from $\mathcal{N}(0,1)$ and set $v_i= \frac{v_i'}{\| v_i' \|}$. We then add noise $\epsilon \sim \mathcal{N}(0,0.2)$. 
The dosage $d \given \xb,t$ follows a Beta distribution with parameter $\alpha$ (default as 2) representing the dosage selection bias. We calculate $t_t = \frac{\alpha-1}{d^*} + 2-\alpha$, with $d^*$ as the optimal dosage for that treatment.

For TCGA(0), we generate $y$ and $d^*$ as follows:
\begin{align}
    y \given \xb,d &= 10(v_1^{\top}\xb + 12d v_3^{\top}\xb  - 12 d^2 v_3^{\top}\xb ), \\
    d^* &= \frac{v_2^{\top}\xb}{2v_3^{\top}\xb}.
\end{align}

For TCGA(1), we generate $y$ and $d^*$
as follows:    $y\given\xb,d = 10((v_1)^{\top}\xb + \sin(\pi (\frac{v_2^{\top}\xb}{v_3^{\top}\xb}d))),
    d^* = \frac{v_3^{\top}\xb}{2 v_2^{\top}\xb}$.

For  TCGA(2),   $ y\given \xb,d = 10(v_1^{\top}\xb + 12d(d-0.75\frac{v_2^{\top}\xb}{v_3^{\top}\xb})^2),
    d^* = 0.25\frac{v_2^{\top}\xb}{v_3^{\top}\xb} \text{ if } \frac{v_2^{\top}\xb}{v_3^{\top}\xb}\ge 1, \text{ else } 1$.

\section{Network Architecture}

\begin{minipage}[!t]{0.4\textwidth}
    \paragraph{Binary Datasets:}
    \our\ is a representation learning-based method. It uses the same architecture as other representation learning-based baselines. The model comprises a Representation Network ($\phi$) with three layers, each consisting of 200 units and ELU (Exponential Linear units) activation functions. Additionally, it includes two $\mu$ Networks, $\mu_0$ (for T=0) and $\mu_1$ (for T=1). These networks follow a similar structure with two layers, featuring 100 units each and ELU activation functions. \our\ estimates the treatment effect $\tau$ as the difference between the two predicted potential outcomes.
\end{minipage}
\hfill
\begin{minipage}[!t]{0.55\textwidth}
    \renewcommand{\arraystretch}{1.2}
        \begin{tabular}{|p{0.9\linewidth}|}
            \arrayrulecolor{gray!50}
            \hline
            \cellcolor{gray!10}\textbf{Network Architecture for Binary Datasets} \\
            \hline
            \textbf{Input Data} \\
            \textbf{Representation Network ($\mathbf{\phi}$)} \\
            \textcolor{blue}{Layer 1: \quad \quad \quad \quad 200 Units, ELU Activation} \\
            \textcolor{blue}{Layer 2:  \quad \quad \quad \quad  200 Units, ELU Activation} \\
            \textcolor{blue}{Embedding Layer: 200 Units, ELU Activation} \\
            \textbf{$\mu_0$ Network (T=0)} \\
            \textcolor{blue}{Layer 1: \quad \quad \quad \quad 100 Units, ELU Activation} \\
            \textcolor{blue}{Layer 2:  \quad \quad \quad \quad100 Units, ELU Activation} \\
            \textcolor{blue}{Output Layer: \quad \;\;1 Unit} \\
            \textbf{$\mu_1$ Network (T=1)} \\
            \textcolor{blue}{Layer 1:  \quad \quad \quad \quad100 Units, ELU Activation} \\
            \textcolor{blue}{Layer 2:  \quad \quad \quad \quad 100 Units, ELU Activation} \\
            \textcolor{blue}{Output Layer: \quad \;\; 1 Unit} \\
            \hline
        \end{tabular}
\end{minipage}

\xhdr{Continuous Datasets}
Here, we consider two architectures namely DRNet, VCNet
\paragraph{DRNet: } We borrow the architecture of the dose-response network \cite{DRNet} which is an extension of the binary treatment effect estimation models to the continuous case. Instead of two heads as in the case of binary treatment, we split the continuous treatments $T \in [0,1]$ into $N = 5$ uniformly spaced bins and assign a separate output head $\mu_k$ for the $k^{th}$ bin, $k \in \{0,\, 1,\, \dots,\, 4\}$. Additionally, for every sample, the treatment $T = t$ is first normalized by subtracting the magnitude of the lower bin edge $\frac{k}{N}$ and then this scalar is concatenated with the input to each linear layer for the corresponding output head $\mu_k$. For \our\ we estimate the treatment effect $\tau$ for any two treatments $t,\, t'$ as the difference between predicted potential outcomes of the corresponding $\mu$ heads.

\paragraph{VCNet: } The varying coefficient network \cite{vcnet} has the same representation network $\phi$ followed by a single output head $\mu(t)$ where the weights are parameterised by the treatment $t$. Let us denote the weights of the $\mu$ head for a treatment $t$ as $\theta(t) \in \RR^d$. Each weight $\theta_i(t)$ is obtained as a smooth function of the treatment $t$. In particular, we apply spline basis expansion on  $t$ using $N = 5$ spline basis functions $\{\alpha_i\}_{i=1}^5$ and then obtain $\theta_i(t)$ for all $i \in [d]$ as $\sum_{k = 1}^{5} a_{ik} \alpha_k(t)$. So, to learn $\mu$ means to learn the coefficients of the linear combination $\{a_{ik}\}$ while keeping the basis functions $\alpha_i(.)$ fixed. For $\{\alpha_i\}_{i=1}^5$, we use the truncated polynomial basis used by \cite{vcnet} with degree $2$ and knots at $\{\frac{1}{3},\, \frac{2}{3}\}$. This results in a basis with the functions $\{1,\, t,\, t^2,\, (\textsc{ReLU}(t - \frac{1}{3}))^2,\, (\textsc{ReLU}(t - \frac{2}{3}))^2 \}$.

In every forward pass, we first initialize the weights of the $\mu(t)$ head for the corresponding treatment $t$ and then pass the representation $\phi(x)$ as input to it. The smooth variation of weights with $t$ allows the network to learn a smooth potential outcome function. \our\ estimates the treatment effect $\tau$ for treatments $t,\, t'$ as $\mu(t')(\phi(x)) - \mu(t)(\phi(x))$

\section{\our\ loss for Twins Dataset with Binary outcome} \label{sec:app:twins}
When dealing with binary outcomes $\mathcal{Y} \in \{0, 1\}$, modeling the difference of observed outcomes for two individuals transforms the problem into a three-way classification task, where the difference labels can take values across $\{-1, 0, +1\}$. To solve this three-way classification task, \our\, however, uses the same architecture as used for other datasets, and converts the estimated potential outcomes into three-way classification logits as shown below. Let

\begin{align}
    \hat{y_0} = \mu(\phi(\xb), t=0) &= P(y = 0|\xb, t=0) \\ \nonumber
    \hat{y_1} = \mu(\phi(\xb), t=1) &= P(y = 0|\xb, t=1)\\ \nonumber
    \hat{y_0'} = \mu(\phi(\xb'), t=0) &= P(y' = 0|\xb', t'=0)\\ \nonumber
    \hat{y_1'} = \mu(\phi(\xb'), t=1) &= P(y' = 0|\xb', t'=1)\\ \nonumber
\end{align}

where we assume a pair of examples $(\xb, t, y, \xb', t', y')$. Recall that by virtue of pairs creation, we have $t \ne t'$. Having estimated all possible potential outcomes for the pairs, we compute the three-way logits as follows:

\begin{align}
    \text{logits}[{-1}] &= \left(\hat{y_0} \cdot (1 - \hat{y_1'}) \right) \cdot (1 - t) + \left(\hat{y_1} \cdot (1-\hat{y_0'}) \right) \cdot t  \quad (\text{for the case when } y=0, y'=1)\\ \nonumber
    \text{logits}[\;\;0] &= \left(\hat{y_0} \cdot \hat{y_1'} \right) \cdot (1 - t) + \left(\hat{y_1} \cdot \hat{y_0'} \right) \cdot t + \left((1 - \hat{y_0}) \cdot (1 - \hat{y_1'}) \right) \cdot (1 - t) + \left((1 - \hat{y_1}) \cdot (1 - \hat{y_0'}) \right) \cdot t \quad (\text{for the case } y=y')\\ \nonumber
    \text{logits}[+1] &= \left((1 - \hat{y_0}) \cdot \hat{y_1'} \right) \cdot (1 - t) + \left((1 - \hat{y_1}) \cdot \hat{y_0'} \right) \cdot t \quad (\text{for the case when } y=1, y'=0)\\ \nonumber
\end{align}

We finally impose a \verb|cross-entropy| loss using the difference label $y-y'$ as the target.

\section{Computational Infrastructure and Default Hyper-parameters}
Our experiments were conducted on a DGX machine equipped with an NVIDIA A100 GPU card, with 80 GB of GPU memory. The DGX machine is powered by an AMD EPYC 7742 64-Core Processor with 256 CPUs, featuring 64 cores per CPU. Our codebase was entirely developed using JAX \cite{jax2018github}, a functional programming-based deep learning library that extends CUDA support for GPU acceleration.

For a fair comparison, we adopt the hyperparameters used in the CATENets benchmark \cite{benchmarking} as is except for the weight associated with the $L_2$ penalty. In each epoch of training, we sample mini-batches of 100 examples (along with their respective pairs for \our) and impose losses on them. We use Adam optimizer with a learning rate set to 1e-4. Training proceeds for a maximum of 1000 epochs, while we perform early stopping based on a 30\% validation set and a patience level of 10. In the case of binary datasets, we use stratified sampling on the treatments to obtain the validation split, while for continuous datasets, we use random sampling.

\section{On the impact of $L_2$ penalty} \label{sec:app:l2}
In this experiment, we analyze the performance of \our\ and baselines under different $L_2$ penalties on the $\phi$ parameters. In our primary results, which are presented in Table \ref{tab:main_bin}, we applied a high $L_2$ penalty scale of 1 to both \our\ and the baselines. Here we present the results when these methods are trained using CATENets default value of 1e-4.

For better exposition, we compare models trained with the default $L_2$ penalty (1e-4) to those trained with a strong penalty (1). To assess the statistical significance of these performance differences, we conduct $p$-tests using models from the main table \ref{tab:main_bin} as baselines. For example, consider the value ``-0.37 (0.01)"  in the first cell of table \ref{tab:app_bin_l2}. It indicates that training the TLearner on IHDP with a $1e-4$ $L_2$ penalty leads to an average error increase of 0.37 across various IHDP seeds, with a $p$-value of 0.01 showing the significance of this error increase.
Throughout the table, a $p$-value below 0.05 suggests that a $L_2$ penalty of 1 is beneficial for the model's performance, while a $p$-value above 0.95 indicates that the strong penalty may hinder performance.
In summary, we draw the following conclusions from Table \ref{tab:app_bin_l2}:

\begin{enumerate}
    \item Negative values for many methods indicate that a stronger penalty is beneficial for these methods.
    \item Additionally, several $p$-values fall below 0.05, indicating statistically significant performance improvements resulting from the stronger $L_2$ penalty.
    \item RLearner, which employs Robinson decomposition to directly model CATE, is the most affected by the strong $L_2$ penalty.
    \item When comparing results across Tables \ref{tab:main_bin} and \ref{tab:app_bin_l2}, our results show that \our\ trained with an $L_2$ penalty of 1 outperforms all baselines trained on either value of $L_2$ penalty.
\end{enumerate}

{\renewcommand{\arraystretch}{1.2}%
\begin{table*}[!h]
    \centering
    \setlength\tabcolsep{3.2pt}
    \resizebox{0.85\textwidth}{!}{
    \begin{tabular}{l|r|r|r|r|r|r}
    \hline
        &  \multicolumn{2}{c|}{IHDP} &  \multicolumn{2}{c|}{ACIC} & \multicolumn{2}{c}{Twins} \\
        \hline
        & PEHE in & PEHE out & PEHE in & PEHE out & PEHE in & PEHE out \\
        \hline \hline
        \first{TLearner \cite{xlearner}} & -0.37 (0.01) & -0.42 (0.06) & -0.58 (0.21) & -0.95 (0.22) & -0.02 (0.00) & -0.03 (0.00) \\
        \first{RLearner \cite{rlearner}} & +0.87 (0.99) & +0.75 (0.96) & +0.87 (0.95) & +0.15 (0.62) & -0.01 (0.00) & -0.02 (0.00) \\
        \first{DRLearner \cite{drlearner}} & -0.39 (0.01) & -0.42 (0.06) & +0.22 (0.64) & -0.03 (0.48) & -0.01 (0.00) & -0.02 (0.00) \\
        \first{XLearner \cite{xlearner}} & -0.24 (0.21) & -0.29 (0.22) & -0.31 (0.29) & -0.75 (0.16) & -0.01 (0.00) & -0.01 (0.00) \\
        \hline
        \second{TARNet \cite{xlearner}} & -0.52 (0.00) & -0.67 (0.00) & -0.30 (0.25) & -0.77 (0.05) & -0.01 (0.01) & -0.01 (0.00) \\
         \second{CFRNet \cite{cfrnet}} & -0.59 (0.00) & -0.68 (0.00) & +0.01 (0.51) & -0.33 (0.30) & 0.00 (0.73) & +0.00 (0.84) \\
        \second{DragonNet \cite{dragonnet}} & -0.51 (0.00) & -0.66 (0.00) & -0.28 (0.26) & -0.74 (0.06) & -0.01 (0.00) & -0.01 (0.00) \\
        \second{FlexTENet \cite{inducbias}} & -0.21 (0.03) & -0.25 (0.10) & +0.67 (0.96) & +1.42 (0.99) & +0.04 (1.00) & +0.05 (1.00) \\
        \hline
        PairNet & -0.98 (0.00) & -1.08 (0.00) & -0.48 (0.09) & -0.89 (0.01) & -0.00 (0.50) & -0.00 (0.49) \\
        \hline

    \end{tabular}}
        \vspace{0.2cm}
       \caption{Differences in Performance with Strong $L_2$ Penalty: We examine the performance differences between models trained under the default $L_2$ penalty setting (1e-4) and those trained with a strong $L_2$ penalty of 1 on the $\phi$ parameters. This table presents the mean differences and corresponding $p$-values in brackets for a one-sided paired t-test conducted using the methods from Table \ref{tab:main_bin} as the baseline. The table illustrates the discrepancy in error compared to our primary results where negative values indicate that the strong $L_2$ penalty outperforms its counterpart. The direct methods are highlighted in \hlc[green]{green} and representation learning-based methods are highlighted in \hlc[yellow]{yellow}.}
       \label{tab:app_bin_l2}
\end{table*}}

\begin{table*}[!h]
    \centering
    \setlength\tabcolsep{3.2pt}
    \resizebox{0.85\textwidth}{!}{
    \begin{tabular}{l|r|r|r|r|r|r}
    \hline
        &  \multicolumn{2}{c|}{IHDP} &  \multicolumn{2}{c|}{News} & \multicolumn{2}{c}{TCGA-0}\\
        \hline
         & PEHE in & PEHE out & PEHE in & PEHE out & PEHE in & PEHE out \\
        \hline
        \hline
        DRNet \cite{DRNet}  & -1.38 (0.00) & -1.43 (0.00) & 0.13 (1.00) & -0.01 (0.29) & 0.06 (0.94) & 0.05 (0.91) \\
        \our\ (DRNet)       & -1.33 (0.00) & -1.91 (0.00) & -0.07 (0.00) & -0.27 (0.00) & -0.05 (0.03) & -0.07 (0.00) \\
        \hline 
        VCNet \cite{vcnet}  & 0.44 (1.00) & 0.17 (0.96) & -0.00 (0.32) & -0.02 (0.04) & -0.01 (0.35) & -0.01 (0.32) \\
        \our\ (VCNet)       & 0.34 (1.00) & 0.12 (0.83) & 0.01 (0.88) & -0.01 (0.09) & -0.00 (0.43) & -0.00 (0.42) \\
        \hline
    \end{tabular}}
    \vspace{0.1cm}
   \caption{Performance of \our\ and baselines on DRNet, VCNet assessed using PEHE error on continuous datasets. We compute the difference in error for two different values of $L_2$ penalty, $1$ and $10^{-4}$. We report differences in mean error and $p$-values in brackets for a one-sided paired t-test conducted with $L_2 = 1$ as the baseline for each model. We see that all the methods benefit significantly from regularizing the $\phi$ parameters.}
   \label{tab:app_cont_l2}
\end{table*}

\section{Results using Shallow Model Architecture}

We performed experiments using the shallow architecture in \cite{benchmarking}. Our main paper featured three layers, each with 200 neurons, in $\phi$, and two layers, each with 200 neurons, in $\mu$. Here, we conducted an experiment using one layer with 200 neurons in $\phi$ and one layer with 100 neurons each in the $\mu$ heads.
We show the PEHE-out for the five binary datasets in Table \ref{tab:shallow}. With this smaller model, \our\ outperforms the baselines by a greater margin than what was reported in our main paper in Table \ref{tab:main_bin}, as indicated by the significance of many p-values in the above table.

\begin{table*}[!h]
    \centering
    \setlength\tabcolsep{3.2pt}
    \resizebox{0.6\textwidth}{!}{
    \begin{tabular}{l|r|r|r|r|r}
    \hline 
        Method & IHDP & ACIC2 & ACIC7 & ACIC26 & Twins\\
        \hline \hline
        TNet & 2.04 (0.02) & 4.14 (0.01) & 3.99 (0.1) & 3.95 (0.06) & \first{0.32 (0.00)} \\
        TARNet & 1.16 (0.2) & 1.78 (0.24) & 3.88 (0.12) & 3.75 (0.05) & 0.33 (0.00) \\
        CFRNet & 1.34 (0.12) & 3.59 (0.2) & 4.56 (0.02) & 4.49 (0.02) & 0.33 (0.00) \\
        RNet & 3.0 (0.00) & 1.18 (0.13) & 5.44 (0.00) & 4.91 (0.05) & \first{0.32 (0.19)} \\
        XNet & 2.18 (0.03) & 3.15 (0.07) & 3.88 (0.11) & 3.55 (0.08) & \first{0.32 (0.02)} \\
        FlexTENet & 1.35 (0.1) & 5.59 (0.02) & 4.54 (0.03) & 6.4 (0.01) & 0.36 (0.00) \\
        DRNet & 1.38 (0.1) & 4.77 (0.11) & 4.51 (0.04) & 3.99 (0.02) & \first{0.32 (0.10)} \\
        DragonNet & 1.15 (0.2) & 1.74 (0.25) & 3.89 (0.12) & 3.78 (0.04) & 0.33 (0.00) \\
        IPW & 1.12 (0.23) & 1.69 (0.24) & 3.43 (0.3) & 3.43 (0.07) & 0.33 (0.00) \\
        NearNeighbor & 1.6 (0.08) & 1.2 (0.14) & 4.3 (0.03) & 4.3 (0.00) & \first{0.32 (0.00)} \\
        PairNet & \first{0.76 (0.00)} & \first{0.7 (0.00)} & \first{3.05 (0.00)} & \first{2.36 (0.00)} & \first{0.32 (0.00)} \\
        \bottomrule
    \end{tabular}}
    \vspace{0.1cm}
   \caption{Performance comparison on shallow model architecture which features one layer with 200 neurons in $\phi$ and one layer with 100 neurons each in the $\mu$ heads. We show the PEHE-out values with $p$-values in brackets. We observe that the performance gains of using pair loss is much more evident on shallow architectures than what was observed on deep architecture shown in Table \ref{tab:main_bin}.}
   \label{tab:shallow}
\end{table*}

\section{Results on Individual ACIC Versions} \label{app:ind_acic}
The main Table \ref{tab:main_bin} featured aggregated results for the ACIC datasets. We compare the performance of \our\ with other baselines on each version of the ACIC dataset in Table \ref{tab:ind_acic}. We see that except on ACIC2, \our\ achieves much better performance on the remaining datasets. 

\begin{table*}[!h]
    \centering
    \setlength\tabcolsep{3.2pt}
    \resizebox{0.45\textwidth}{!}{
    \begin{tabular}{l|r|r|r}
    \hline 
        Method & ACIC2 & ACIC7 & ACIC26 \\
        \hline \hline
        TNet & 3.39(0.01) & 3.50(0.1) & 6.00(0.14) \\
        TARNet & \first{1.15(0.68)} & 3.49(0.09) & 3.5(0.22) \\
        CFRNet & 2.12(0.36) & 4.12(0.01) & 4.12(0.06) \\
        RNet & 1.26(0.71) & 5.30(0.00) & 5.25(0.01) \\
        XNet & 1.80(0.39) & 3.51(0.07) & 4.60(0.15) \\
        FlexTENet & 4.89(0.00) & 4.62(0.00) & 6.60(0.00) \\
        DRNet & 2.50-(0.28) & 3.85(0.03) & 3.65(0.16) \\
        DragonNet & \first{1.15(0.68)} & 3.49(0.09) & 3.52(0.21) \\
        IPW & 1.35(0.59) & \first{2.85(0.53)} & 3.5(0.24) \\
        NearNeighbor & 1.24(0.72) & 4.09(0.00) & 4.10(0.06) \\
        PairNet & 1.56(0.00) & 2.89(0.00) & \first{2.95(0.00)} \\
        \bottomrule
    \end{tabular}}
    \vspace{0.1cm}
   \caption{Performance comparison on individual ACIC seeds.}
   \label{tab:ind_acic}
\end{table*}

\section{Results including PEHE in for continuous datasets} \label{sec:app:cont_pehe_in}
We present the results for PEHE in errors in table \ref{tab:app_cont} for completeness. We observe that both PEHE in and PEHE out errors exhibit similar trends across the datasets.
\begin{table*}[!h]
    \centering
    \setlength\tabcolsep{3.2pt}
    \resizebox{1.0\textwidth}{!}{
    \begin{tabular}{l|r|r|r|r|r|r|r|r|r|r}
    \hline
        &  \multicolumn{2}{c|}{IHDP} &  \multicolumn{2}{c|}{News} & \multicolumn{2}{c|}{TCGA-0}& \multicolumn{2}{c|}{TCGA-1}& \multicolumn{2}{c}{TCGA-2} \\
        \hline
         & PEHE in & PEHE out & PEHE in & PEHE out & PEHE in & PEHE out & PEHE in & PEHE out & PEHE in & PEHE out \\
        \hline
        \hline
        DRNet \cite{DRNet} & 2.29 (0.00) & 2.45 (0.00) & 1.42 (0.00) & 1.42 (0.00) & 0.34 (0.01) & 0.34 (0.01) & 0.26 (0.07) & 0.26 (0.08) & 0.24 (0.02) & 0.24 (0.02)\\
        \our\ (DRNet) & $\first{2.03 (0.00)}$ & $\first{ 2.27 (0.00) }$& $\first{1.21 (0.00) }$& $\first{1.32 (0.00)}$ & $\first{0.26 (0.00)}$ & $\first{0.26 (0.00)}$ & $\first{0.22 (0.00)}$ & $\first{0.22 (0.00)}$ & $\first{0.21 (0.00)}$ & $\first{0.21 (0.00)}$ \\
        \hline 
        VCNet \cite{vcnet}& 1.59 (0.00) & 1.73 (0.02) & $\first{ 1.13 (1.00)}$ & $\first{1.24 (1.00)}$ & 0.25 (0.57) & 0.25 (0.57) & 0.22 (0.45) & 0.22 (0.45) & 0.21 (0.45) & 0.21 (0.46) \\
        \our\ (VCNet)& $\first{1.43 (0.00)}$ & $\first{1.57 (0.00)}$ & 1.14 (0.00) & 1.26 (0.00) & 0.25 (0.00) & 0.25 (0.00) & 0.22 (0.00) & 0.22 (0.00) & 0.21 (0.00) & 0.21 (0.00) \\
        \hline
    \end{tabular}}
    \vspace{0.1cm}
   \caption{Performance of \our\ losses on DRNet, VCNet assessed using PEHE error on continuous datasets. We report mean and $p$-values within brackets for a one-sided paired t-test conducted with \our\ as the baseline.}
   \label{tab:app_cont}
\end{table*}

{\renewcommand{\arraystretch}{1.3}%
\begin{table}[!h]
    \centering
    \setlength\tabcolsep{3.2pt}
    \resizebox{0.7\textwidth}{!}{\resizebox{\textwidth}{!}{\begin{tabular}{c|c|c|c|c|c|c}
        \hline
        & \multicolumn{3}{c|}{TCGA0} & \multicolumn{3}{c}{TCGA2} \\
        \cline{2-7}
         Trn. Size & $5410$ & $1352$ & $541$ & $5410$ & $1352$ & $541$ \\
        \hline
        \hline
        Factual & 0.25 (0.51) & 0.28 (0.14) & 0.43 (0.02) & 0.45 (0.52) & 0.46 (0.44) & 0.58 (0.13)\\
     \our\ &0.26 (0.00) & \first{0.25 (0.00)} & \first{0.27 (0.00)} & 0.45 (0.00) & 0.46 (0.00) & \first{0.49 (0.00)} \\
        \hline
    \end{tabular}}}
    \vspace{0.1cm}
   \caption{PEHE out error for VCNet}
   \label{tab:app:vcnet}
\end{table}

{\renewcommand{\arraystretch}{}%
\begin{table}[!h]
    \centering
    \setlength\tabcolsep{3.2pt}
    \resizebox{0.4\textwidth}{!}{
    \begin{tabular}{l|r|r|r}
        \hline
         ~ &  \multicolumn{1}{c|}{Random} & \multicolumn{1}{c|}{\our's $\phi$}& \multicolumn{1}{c}{\third{$\embedfct$}}\\
    \hline
    \hline
    IHDP & 0.45 (0.42) & 0.46 (0.52) & 0.45 (0.00) \\
    ACIC-2 & 1.26 (0.88) & 1.53 (0.84) & 0.69 (0.00) \\
    \hline
    IHDP (DRNet) & 2.28 (0.57) & 2.27 (0.60) & 2.29 (0.00) \\
    IHDP (VCnet) & 1.57 (0.49) & 1.56 (0.51) & 1.57 (0.00)\\
    \hline
    \end{tabular}}
    \vspace{0.1cm}
     \caption{PEHE out error for various schemes of deriving pairs both the binary and continuous treatment settings assessed across five seeds. Overall we observe that \our\ is not sensitive to the different choices, except in the ACIC-2 dataset when the $\embedfct$ performs the best.}
     \label{tab:abl:phifct}
\end{table}}

\section{Impact of Embedding Function for Pairing}\label{app:sec:phifct}
\our\ uses embeddings $\embedfct(\xb)$ to calculate distances when creating pairs. We evaluate the impact of different embedding functions on performance for both real and synthetic data and show the importance of a good distance measure.
We consider three variants of embedding function:
(a) Random: pairs are selected arbitrarily, 
(b) Factual $\embedfct$: uses embeddings from a pre-trained model, trained solely using the factual loss.
(c) \our's $\phi$: pairs are generated at each epoch using distances computed on \our's representation network $\phi$, trained up to that epoch.

The results are shown in table \ref{tab:abl:phifct}. For IHDP, we see that \our\ exhibits resilience to different choices for the embedding function. Similar trends are seen for most other datasets (not shown here). However, for the second version of the the ACIC dataset, we notice that the use of $\embedfct$ leads to a significant reduction in error over random selection. This indicates that finding a good distance measure for samples can boost the performance of our algorithm but even for a poor measure we do not do worse than random selection in most cases.

{\renewcommand{\arraystretch}{1.2}
\begin{table}[]
    \centering
    \resizebox{0.6\textwidth}{!}{\begin{tabular}{c|c|c|c|c|c}
    \hline
    & \multicolumn{4}{c|}{\our} &\\
    \cline{2-5}
     & A & B & A+B & Random & Factual \\
    \hline
    \hline
    PEHE in & \first{0.98 (0.00)} & 1.01 (0.48) & 1.00 (0.49) & 1.03 (0.47) & 1.03 (0.47) \\
    PEHE out  &  \first{0.58 (0.00)}& 0.61 (0.47)& 0.60 (0.48)& 0.64 (0.44) & 0.61 (0.47) \\
    \hline
\end{tabular}}
    \caption{PEHE error on synthetic data under different pairing strategies for \our. Strategy A which relies on relevant covariates for pair construction outperforms other approaches highlighting the importance of a good embedding function.}
    \label{tab:abl:phifct_synthetic}
\end{table}}

\section{Correct Pairing Boosts \our}
We further conduct a toy experiment to bring out the point that finding correct pairs boosts \our's performance while arbitrary pairing does not hurt beyond a factual model. We considered i.i.d. Gaussian covariates $\xb \in \RR^{10}$. We assumed $\mu_0$, $\mu_1$ to be 3-degree polynomials such that their difference $\tau$ is of degree-2. Suppose $\mu_0, \mu_1$ solely depends on the first 5 covariates, we test four distance measure variants for pair computation in \our: 
(1) A - computes distances using $\xb[0:5]$
(2) B - uses irrelevant covariates $\xb[5:10]$
(3) A+B uses all covariates
(4) Random - arbitrary pairing

As expected, \our-A outperforms others, emphasizing the significance of nearby pairing and a robust distance measure. This observation aligns with \our's $\phi_{\text{fct}}$, extracting predictive features for $\mu$, analogous to $\xb[0:5]$ here.

To avoid over-representation of certain examples while pairing, we also considered techniques like Optimal Transport, but they did not perform much better than the presented approaches. 

\section{Details on the Code}

We have uploaded the code with our supplementary material. The code is accompanied by a \verb|README.MD| that specifies the installation instructions, and directions on how to run the code.

\section{Limitations}
These are the main limitations of our work:
\newcommand{\SubItem}[1]{
    {\setlength\itemindent{15pt} \item[-] #1}
}

\begin{enumerate}
    \item \our\ imposes losses only on the factual observed outcomes. Therefore, even for the cases where distant examples are paired, its performance is not expected to degrade beyond models trained solely on factual losses. We face the following challenges in finding good pairs:
    \SubItem{
    There should be an adequate number of proximate pairs in the observational dataset for \our\ to find them.}
    
    \SubItem{The requirement for proximal pairs with opposite treatments in the training data is mitigated by the overlap assumption that is generally made for causal inference tasks. Overlap states that $P(t|X) > 0\;\; \text{ }\forall\text{ }t \in T$. However, with finite observational datasets, this assumption may not hold for certain covariates, resulting in the pairing of distant ones.}

    \item Another limitation arises when opting to apply losses to more than one pair ($\text{num}_{z'}>1$) for each observed sample. This results in a computational time increase, albeit only linearly proportional to $\text{num}_{z'}$. Furthermore, generating pairs may be time-consuming, especially when dealing with large datasets. Nevertheless, efficient techniques such as FAISS \cite{faiss} can be employed to perform the pairing efficiently.
\end{enumerate}

\end{document}